\documentclass[sigconf]{acmart}
\RequirePackage{float}  
\RequirePackage{subfigure}  
\usepackage{multirow}

\AtBeginDocument{%
  \providecommand\BibTeX{{%
    \normalfont B\kern-0.5em{\scshape i\kern-0.25em b}\kern-0.8em\TeX}}}

\copyrightyear{2024}
\acmYear{2024}
\setcopyright{rightsretained}
\acmConference[MM '24]{Proceedings of the 32nd ACM International Conference on Multimedia}{October 28-November 1, 2024}{Melbourne, VIC, Australia}
\acmBooktitle{Proceedings of the 32nd ACM International Conference on Multimedia (MM '24), October 28-November 1, 2024, Melbourne, VIC, Australia}
\acmDOI{10.1145/3664647.3680804}
\acmISBN{979-8-4007-0686-8/24/10}

\makeatletter
\gdef\@copyrightpermission{
  \begin{minipage}{0.3\columnwidth}
   \href{https://creativecommons.org/licenses/by-nc/4.0/}{\includegraphics[width=0.90\textwidth]{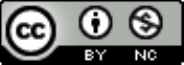}}
  \end{minipage}\hfill
  \begin{minipage}{0.7\columnwidth}
   \href{https://creativecommons.org/licenses/by-nc/4.0/}{This work is licensed under a Creative Commons Attribution-NonCommercial International 4.0 License.}
  \end{minipage}
  \vspace{5pt}
}
\makeatother

\begin{document}

\title{Spatiotemporal Graph Guided Multi-modal Network for Livestreaming Product Retrieval}


\author{Xiaowan Hu}\authornotemark[1]
\affiliation{%
  \institution{The Shenzhen International Graduate School, Tsinghua University}
  \city{Shenzhen}
  \country{China}}
\email{huxw22@mails.tsinghua.edu.cn}

\author{Yiyi Chen}\authornote{Equal contribution. This work is done during an internship at Kuaishou Technology.}
\affiliation{%
  \institution{Kuaishou Technology}
  \city{Beijing}
  \country{China}}
\email{chenyiyibupt@gmail.com}

\author{Yan Li}
\affiliation{%
  \institution{Kuaishou Technology}
  \city{Beijing}
  \country{China}}
\email{yan.li@cripac.ia.ac.cn}

\author{Minquan Wang}
\affiliation{%
  \institution{Kuaishou Technology}
  \city{Beijing}
  \country{China}}
\email{wangminquan@kuaishou.com}

\author{Haoqian Wang}
\affiliation{%
  \institution{The Shenzhen International Graduate School, Tsinghua University}
  \city{Shenzhen}
  \country{China}}
\email{wanghaoqian@tsinghua.edu.cn}

\author{Quan Chen}
\affiliation{%
  \institution{Kuaishou Technology}
  \city{Beijing}
  \country{China}}
\email{chenquan06@kuaishou.com}

\author{Han Li}
\affiliation{%
  \institution{Kuaishou Technology}
  \city{Beijing}
  \country{China}}
\email{lihan08@kuaishou.com}

\author{Peng Jiang}
\affiliation{%
  \institution{Kuaishou Technology}
  \city{Beijing}
  \country{China}}
\email{jiangpeng@kuaishou.com}

\renewcommand{\shortauthors}{Xiaowan Hu et al.}

\begin{abstract}
With the rapid expansion of e-commerce, more consumers have become accustomed to making purchases via livestreaming. Accurately identifying the products being sold by salespeople, i.e., livestreaming product retrieval (LPR), poses a fundamental and daunting challenge. The LPR task encompasses three primary dilemmas in real-world scenarios: 1) the recognition of intended products from distractor products present in the background; 2) the video-image heterogeneity that the appearance of products showcased in live streams often deviates substantially from standardized product images in stores; 3) there are numerous confusing products with subtle visual nuances in the shop. To tackle these challenges, we propose the Spatiotemporal Graphing Multi-modal Network (SGMN). First, we employ a text-guided attention mechanism that leverages the spoken content of salespeople to guide the model to focus toward intended products, emphasizing their salience over cluttered background products. Second, a long-range spatiotemporal graph network is further designed to achieve both instance-level interaction and frame-level matching, solving the misalignment caused by video-image heterogeneity. Third, we propose a multi-modal hard example mining, assisting the model in distinguishing highly similar products with fine-grained features across the video-image-text domain. Through extensive quantitative and qualitative experiments, we demonstrate the superior performance of our proposed SGMN model, surpassing the state-of-the-art methods by a substantial margin. The code is available at \url{https://github.com/Huxiaowan/SGMN}.
\end{abstract}





\maketitle

\section{Introduction}
Benefiting from the convenience of e-commerce, shopping online become increasingly popular in recent years. The livestreaming product retrieval (LPR) plays a crucial role in accurately matching products shown in live clips with those available in online shops, ensuring a seamless shopping experience and excellent marketing ability~\cite{hadi2015buy, liu2016deepfashion, ge2019deepfashion2}. Even though various customer-to-shop retrieval methods have great progress~\cite{huang2015cross, cheng2017video2shop, guan2022bi, godi2022movingfashion,zhou2023contrastive}, the cross-domain livestreaming-to-shop problem in LPR has few studies yet. Some unresolved challenges are still rooted in real-world applications. 

\begin{figure}[t!] 
    \centering 
    \subfigtopskip=1pt 
    \subfigbottomskip=-1.7pt
    \subfigcapskip=-3pt 
    \subfigure[Cluttered and similar background products.]{
        \label{product}
        \centering
        \includegraphics[width=80mm]{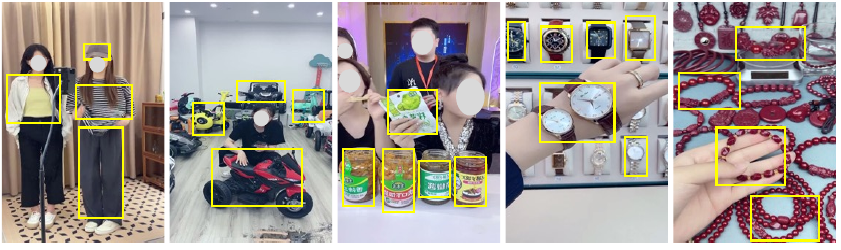}}
    \vspace{1mm} 
    \subfigure[Intended product retrieval with a large appearance variations.]{
        \label{vis_graph}
        \centering
        \includegraphics[width=81.5mm]{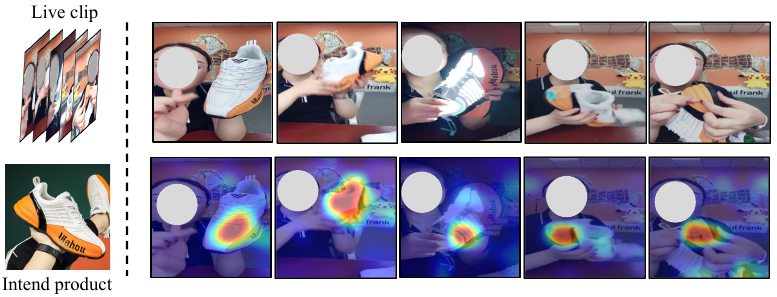}} 
    \subfigure[Intra-domain and inter-domain spatiotemporal graphs.]{
        \label{st_graph}
        \centering
        \includegraphics[width=82mm]{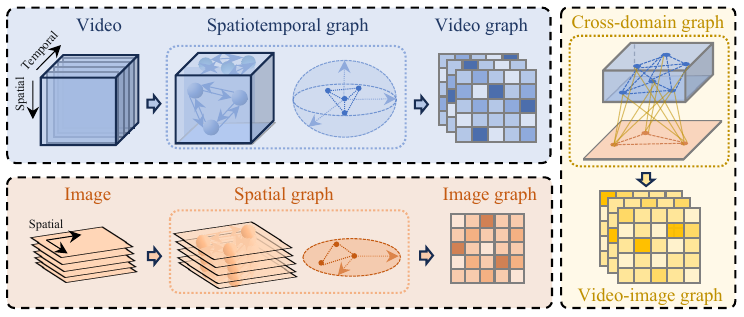}}
    \vspace{-2.5mm}  
    \caption{Representative examples of cluttered and similar background products (a) or large appearance variations such as occlusion, motion, and illumination (b). The intra-domain and inter-domain graphs (c) between videos and images to enhance the spatiotemporal frame-level interaction.}
    \label{overall_map}
\vspace{-5.8mm}    
\end{figure}

First of all, accurately discerning the intended products offered by salespersons during livestreaming is profoundly challenging. It is common practice for salespersons to showcase multiple products but focus on promoting a specific product at a given time, defined as intended product. Some methods have attempted to incorporate an additional product detection module~\cite{zhao2021dress, godi2022movingfashion}, but they will incur high annotation costs and computational complexity, and also fail to completely eliminate ambiguity of intent. Sometimes, the detection of multiple foreground boxes introduces more uncertainty and erroneous guidance, as shown in Fig.~\ref{product}.


Secondly, the heterogeneity between realistic livestreaming videos and product images further exacerbates the difficulty of cross-domain retrieval. As depicted in Fig.~\ref{vis_graph}, livestreaming introduces variations in viewpoint, illumination, and occlusions, resulting in significant appearance disparities between products in livestreaming and in online store. It is very common in live scenarios, where methods focused on the instance-level matching of entire clips will fail~\cite{cheng2017video2shop,godi2022movingfashion,yang2023crossview}.
Coarse-grained instance-level matching makes it hard to track spatial structural deformations. The inevitable motion and occlusions will render products visually obscured or degraded during specific time slots, making frame-level sequential matching with temporal consistency necessary. Moreover, 
Large domain discrepancy leads to misalignment of intra-domain features. Existing cross-domain retrieval methods~\cite{cheng2017video2shop, fang2021clip2video, liu2021cma, xu2020proposal, ma2022x} often treat different domains equally while ignoring spatiotemporal relations within and between domains,  further reducing retrieval accuracy.

Another dilemma lies in the abundance of highly similar products in online stores, necessitating that models excel at learning fine-grained representations and discriminating subtle distinctions among similar-looking products. As shown in Fig.~\ref{product}, there exists a plethora of indistinguishable bracelets. Some methods based on pre-trained CLIP attempt to learn classifiable visual embeddings from large-scale video-text embeddings~\cite{sun2019videobert, li2019visualbert, chen2020uniter,fang2021clip2video}. However, generic features often fail to capture subtle product characteristics. 


To tackle the real-world challenges mentioned above, we propose the Spatiotemporal Graphing Multi-modal Network (SGMN) for LPR. Our model consists of three key components: 1) We leverage the verbal explanations provided by salespersons in livestreams, which often contain explicit information about the intended products. The texts from Automatic Speech Recognition (ASR)  transcriptions and image titles guide the model to focus on products highly relevant to the verbal context, mitigating distractions from background clutter items. 2) We design a Graph-based Cross-domain Interaction (GCI) module to capture spatiotemporal correlations between videos and images. It is the first exploration of using sequence-to-sequence graph learning to model and enhance cross-domain temporal consistency and spatial correlation. As shown in Fig.~\ref{st_graph}, we establish the intra-domain and inter-domain connection graphs between video and image synchronously. Benefiting from the frame-level connectivity, the model can still accurately localize regions of intended products that encounter appearance distortion due to occlusion, motion, and brightness changes in Fig.~\ref{vis_graph}. 3) The Selective Multi-modal Fusion (SMF) module is proposed to assist the model in distinguishing highly similar products. This mechanism selects the top-K hard examples in global ranking, and then fuses their visual and textual representations for implicitly recalibrating ranking and discerning semantic heterogeneity. The fine-grained alignment across the video-image-text domain demonstrates robust potential. Notably, our method adheres to the principle of independently encoding multi-modal inputs during inference, striking a balance between efficiency and accuracy. 

\noindent  Our contributions can be summarized as follows:
\vspace{-1mm}
\begin{itemize}
	\item We introduce the textual information in the detector-free retrieval framework to guide model attention on the intended product. The comprehensive yet unified representation learned from the cross-modal alignment space of video-image-text solves the practical dilemma of LPR.
	\item The spatiotemporal graph learning is first explored for capturing sequential relations, alleviating inter-domain misalignments in both spatial and temporal dimensions.
        \item We selectively fuse multi-modal features to enhance fine-grained representations of hard samples, distinguishing products with subtle visual differences.
	\item Our method achieves the best performance on the large-scale benchmark dataset. Extensive quantitative and qualitative experiments prove the superiority of SGMN.
\end{itemize}

\vspace{-2.5mm}
\section{Related works}
\subsection{Livestreaming Product Retrieval}
Before the livestreaming became popular, researchers paid more attention to the customer-to-shop clothes retrieval tasks, such as fashion retrieval~\cite{liu2016deepfashion, hadi2015buy, ji2017cross, gajic2018cross, ge2019deepfashion2}. Motivated by video-to-shop retrieval, some works are dedicated to providing pair-matching solutions between image and video features, such as DPRNet~\cite{zhao2021dress} and SEAM~\cite{godi2022movingfashion}. Such methods adopt a two-stage framework combining detection and retrieval, localizing the products in the video before performing the global similarity match. AsymNet~\cite{cheng2017video2shop} uses a single-stage network that removes object detection to reduce model complexity. RICE~\cite{yang2023crossview} proposes a single-stage network without object detection to reduce model complexity. However, these methods have yet to effectively leverage the textual modality to identify the intended products. In our work, we make the first attempt at enhancing visual representations of intended products using the textual features in a one-stage framework.
\begin{figure*}[t]
    \centering
    \includegraphics[width=1\linewidth]{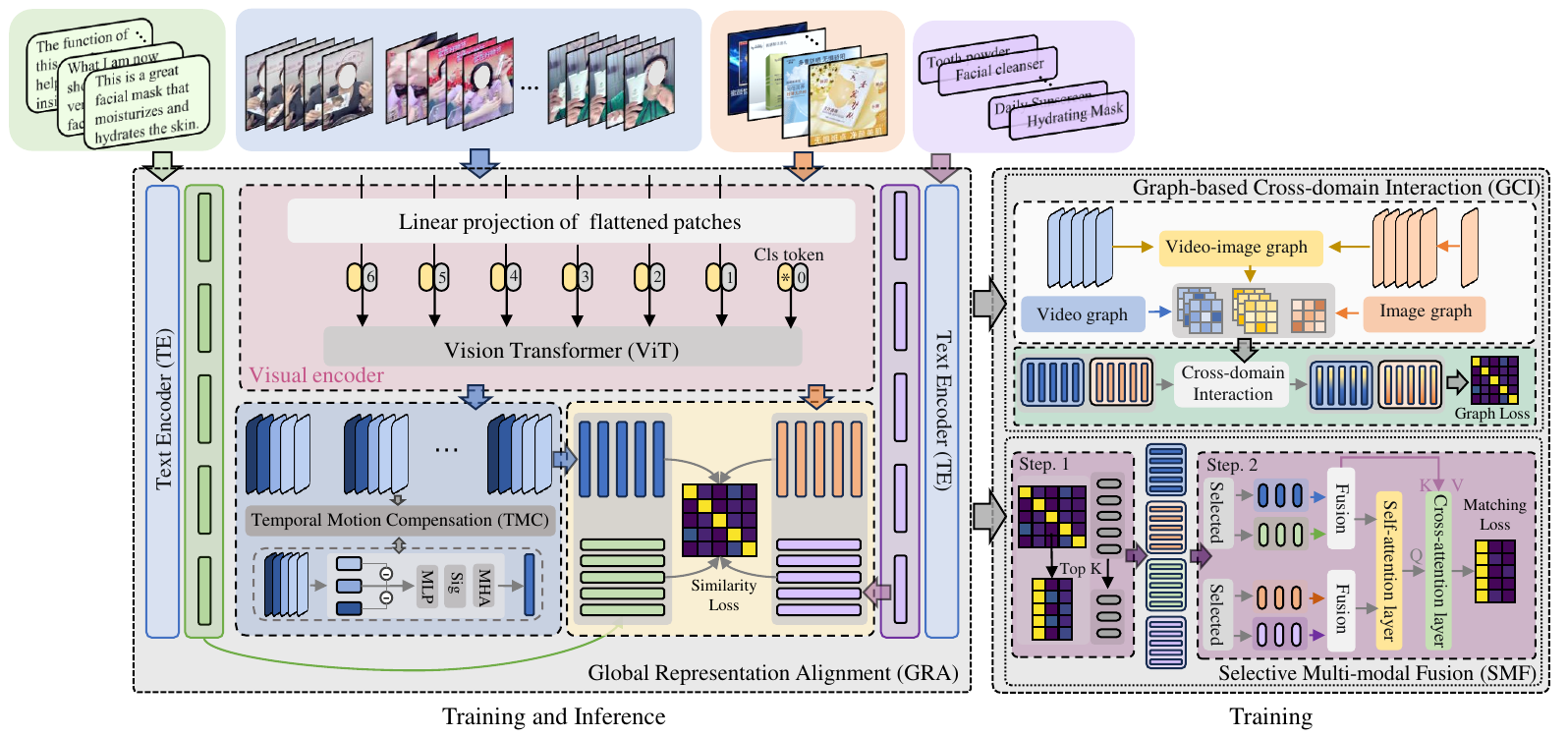}
    \vspace{-7.5mm} 
    \caption{The architecture of the proposed SGMN. The inputs are live video clips, text from video ASR, product images, and product titles. Paired image-video and ASR-title representations in GRA module are independently encoded and weighted for global similarity. The GCI module (Top right) constructs the video graph, image graph, and video-image graph for cross-domain spatiotemporal relation learning. The SMF module (Bottom Right) selects hard examples and fuses multi-modal features for distinguishing mining. Only the GRA module is used for inference, while the GCI and SMF modules are applied for training.}
    \label{pipeline}
\vspace{-3mm} 
\end{figure*}

\vspace{-2.5mm}
\subsection{Cross-domain Retrieval and Interaction}
Existing cross-domain retrieval methods usually learn unified representations across different domains. Some methods feed concatenated visual and textual features into classifiers to predict image-to-text matching performance~\cite{su2019vl, sun2019videobert, li2019visualbert, chen2020uniter}. CLIP2Video~\cite{fang2021clip2video} transfers knowledge from pretrained CLIP to learn video-to-text features. Recent methods with high inference performance use independent encoders to extract global features and compute cross-domain similarity~\cite{lu2019vilbert, tan2019lxmert, radford2021learning, zhou2022conditional, ma2022x}. Methods for cross-domain interaction are also explored. SCAN~\cite{lee2018stacked} uses an attention mechanism to fuse across domains, and IMRAM~\cite{chen2020imram} applies an iterative network for interaction. The multi-grained matching mechanism\cite{ma2022x} shows limited performance in capturing long-range temporal dependency. Our work employs a spatiotemporal graph to enhance the cross-domain consistency across the video-image-text domain.

\vspace{-2.5mm}
\subsection{Reason Graph Learning}
The graph reasoning network (GCN) has proven effective due to the powerful expressive capability of graph structures\cite{zhao2014affective,velivckovic2017graph, chen2019graph}. In cross-domain retrieval tasks, using graph reasoning networks~\cite{ kuang2019fashion, zhou2020graph,ji2022heterogeneous} to learn the relationship between different domains is a common practice. VSRN~\cite{li2019visual} models the local visual features of key objects semantically. CGMN~\cite{cheng2022cross} and GSMN~\cite{liu2020graph} learn image-text matching using a cross-domain graph matching network. DSRAN \cite{wen2020learning} adopts the graph attention. 
Wang ~\textit{et al.} used a coarse-to-Fine graph network for video-text retrieval~\cite{wang2020learning}. The HREM~\cite{fu2023learning} learns semantic relationships among image and text via a hierarchical graph relation model. These methods have shown the advantages of graph learning in modeling spatial correlation, but its potential in capturing temporal information within video sequences has yet to be explored. Our work has made preliminary attempts and validations in spatiotemporal sequence-level graphing.

\vspace{-1.5mm}
\section{Method}
Our SGMN consists of three parts: the Global Representation Alignment (GRA), the Graph-based Cross-domain Interaction (GCI), and the Selective Multi-modal Fusion (SMF), as illustrated in Fig.~\ref{pipeline}. Further implementation details are elucidated in subsequent sections.

\vspace{-2mm}
\subsection{Global Representation Alignment}
\textbf{Image Encoder.} We use the pretrained ViT-B/32 model from CLIP~\cite{luo2022clip4clip} as the image encoder. We define a batch of $N$ images as ${I} \in \mathbb{R}^{N\times H\times W}$, and the encoders extract non-overlapping image patches with the size of $p\times p$. Then we perform a linear projection to project these flattened patches into 1D markers as shown in Fig.~\ref{pipeline}. We then use the Multi-Head Attention (MHA) mechanism to interact with each patch of the image for the global aggregated features $I_{cls}$ and patch-level hidden embedding $I_{hid}$:
\vspace{-1mm}  
\begin{equation}
\{I_{cls}, I_{hid}\} = \operatorname{\mathcal{F}_{MHA}}({{\mathcal Q}_I}, {{\mathcal K}_I},{\mathcal V_I}),
\vspace{-1mm} 
\end{equation} 
where ${\mathcal Q}$, ${\mathcal K}$, and ${\mathcal V}$ represent the query, key, and value. The $I_{cls}\in \mathbb{R}^{N\times D}$ and $D$ is the embedding dimension.

\noindent\textbf{Video Encoder.} The live streams are divided into clips ${V} \in \mathbb{R}^{N\times L\times H\times W}$, where $L$ is the length of video frames. We share parameters between the image encoder and the video encoder for global representation alignment, so the sequential features of video clips are encoded as:
\vspace{-1mm} 
\begin{equation}
\{V_{cls}, V_{hid}\} = \operatorname{\mathcal{F}_{MHA}}({{\mathcal Q}_V}, {{\mathcal K}_V},{\mathcal V_V}).
\vspace{-1mm} 
\end{equation}
We also adopt the cls token $V_{cls}\in \mathbb{R}^{N\times L \times D}$ as the global representation of video clips. To enhance inter-frame alignment inside videos and extract correlations within each frame, we use a Temporal Motion Compensation (TMC) module. TMC adds the displacement of inter-frame actions to the video sequence to simulate motion changes as $\Delta{f_t} = {f^t}-{f^{t-1}}$, where ${f^{t-1}}$ and ${f^{t}}$ represent two adjacent frames. We then encode the motion of consecutive frames and insert them as a motion compensation token for enhancing differential-level attention. The global video representation after motion-compensated enhancement is:
\begin{equation}
V_{visual} = \operatorname{\mathcal{F}_{TMC}}(V_{cls}~|~\left\{\Delta{f_1}, \Delta{f_2}, \ldots, \Delta{f_t}\right\}).
\end{equation}

\noindent\textbf{Text Encoder.} ChineseCLIP~\cite{yang2022chinese} is used to extract initial textual representations for video automatic speech recognition (ASR) $T_{asr}$ and image titles $T_{item}$. To keep the model lightweight, we pre-trained the ChineseCLIP model and fixed the parameters in model training. Since the speech information from the salesperson always contains redundant product-independent information in the real-world livestreaming, we add a filter layer $\operatorname{\mathcal{F}_{filter}}$ to eliminate irrelevant units while retaining key information. 
\begin{equation}
\begin{aligned}
V_{text} = \operatorname{\mathcal{F}_{filter}}(\operatorname{\mathcal{F}_{clip}}(T_{asr})),\\
I_{text} = \operatorname{\mathcal{F}_{filter}}(\operatorname{\mathcal{F}_{clip}}(T_{item})),
\end{aligned}
\end{equation}
where the $\operatorname{\mathcal{F}_{clip}}(\cdot)$ is the function of ChineseCLIP.

\noindent\textbf{Similarity Loss.} We use the triplet loss $\mathcal{T}(\cdot)$~\cite{faghri2017vse} to measure the similarity of visual and textual embedding. The optimization goal of GRA is defined as the similarity loss $\mathcal{L}_{s}$ between video-to-image and text-to-text representations, and $\lambda$ is the loss weighting factor.
\begin{equation}
\mathcal{L}_{s} = \mathcal{T}(V_{visual}, I_{visual})+ \lambda \mathcal{T}(V_{text}, I_{text}).
\label{lamda}
\end{equation}
\subsection{Graph-based Cross-domain Interaction}
\begin{figure}[t]
    \centering
    \includegraphics[width=81mm]{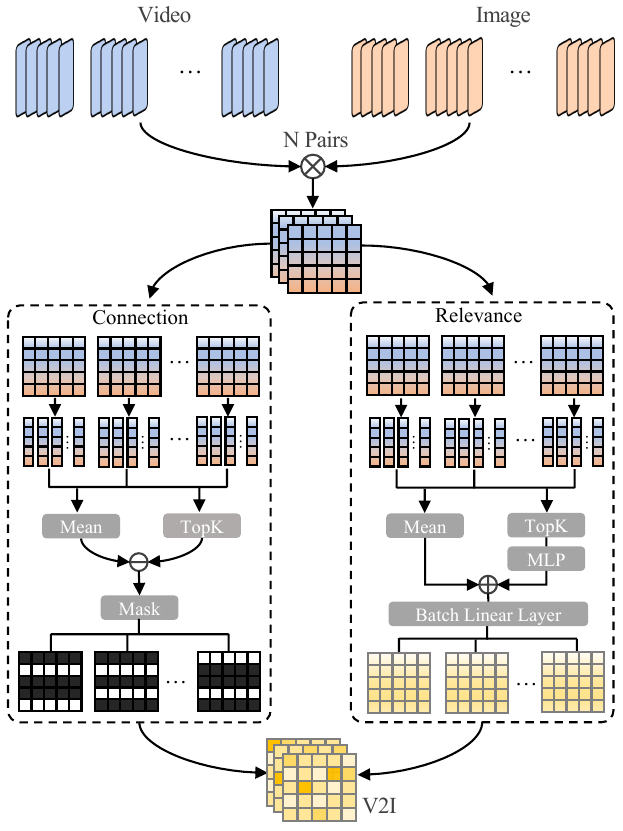}
    \vspace{-3mm} 
    \caption{Graph construction of a cross-domain connection between video and image sequences within a batch.}
    \label{graph}
\vspace{-5mm} 
\end{figure}

\textbf{Graphs Reconstruction.} We build a video graph, image graph, and video-image graph using global representations, as shown in Fig.~\ref{pipeline}. These spatiotemporal graphs can capture temporal consistency and spatial relations synchronously. First, for a batch with $N$ video-image pairs, the sequence length of the video is $L$, so we can obtain the global video features as ${S_V}={V_{visual}}\in \mathbb{R}^{N\times L\times D}$. Then we stack images into sequences of the same length as video clips for the sequence-to-sequence alignment, and the stacked global image features as $S_I = Stack(I_{visual})\in \mathbb{R}^{N\times L\times D}$. The construction of the inter-domain spatiotemporal graph is shown in Fig.~\ref{graph}. The video-to-image similarity matrix $\mathcal{H}{_{V2I}}\in \mathbb{R}^{N\times L\times L}$ is defined as:
\begin{equation}
\begin{aligned}
& \mathcal{H}{_{V2I}}= S_VS_I^T.
\end{aligned} 
\end{equation}

However, fully connecting each $<$frame, image$>$ pair while ignoring the relative importance is suboptimal~\cite{seidenschwarz2021learning}. Thus we define two intra-domain relation matrices and two inter-domain relation matrices, which are video spatiotemporal graph $\boldsymbol{G}_{V2V}$, image spatial graph $\boldsymbol{G}_{I2I}$ and video-image cross-domain graph $\boldsymbol{G}_{V2I}$ and $\boldsymbol{G}_{I2V}$. Then we build the entire spatiotemporal graph with intra-domain relations and cross-domain association as follows:
\begin{equation}
\boldsymbol{G}=\left[\begin{array}{ll}
\boldsymbol{G}_{I2I} & \boldsymbol{G}_{I2V} \\
\boldsymbol{G}_{V2I} & \boldsymbol{G}_{V2V}
\end{array}\right]
\end{equation}

Given $N$ image-video pairs, the size of four relation matrices are $\{{\boldsymbol{G}_{V2V}}, {\boldsymbol{G}_{V2I}, {\boldsymbol{G}_{I2I}}, \boldsymbol{G}_{I2V}}\} \in \mathbb{R}^{N\times L\times L}$. The matrices $\boldsymbol{G}_{V2V}$ and $\boldsymbol{G}_{I2I}$ represent the instance-level correlation of images or videos, while the matrices $\boldsymbol{G}_{V2I}$ and $\boldsymbol{G}_{I2V}$ represents the frame-level correlation between each image and each video frame. All nodes and edges in graphs independently and jointly represent objects and relations of multiple domains. Then we refer to the frame-by-frame matched scores to filter out irrelevant connectivity and define two semantic properties: \textit{Connection} and \textit{Relevance}.

\textbf{i) Connection.} To build an efficient correlation graph and preserve useful spatiotemporal connections, we follow an advanced fine-grained matching scheme~\cite{lee2018stacked}. The connection relationships between each video frame and the image depend on the matching relationships of multiple nodes. 

As shown in Fig.~\ref{graph}, given the frame-image similarity matrix, each column vector represents the matching relationships between each video frame and a certain image. We first compute the mean of the similarity and then measure the top-K matching pairs. We calculate the difference between the mean value and top-K values as:
\begin{equation}
\mathcal{C}_i = \operatorname{Mean}(\mathcal{H}^i_{V2I})-\operatorname{topK}(\mathcal{H}^i_{V2I}, K),\\
\label{eq8}
\end{equation}
where $i \in [0,L)$ and $\mathcal{H}^i_{V2I}$ is the $i$-th column vector and $K$ is set as $L/2$. Finally, the connection property between the frame and the image is determined based on the logical values of $\mathcal{C}$. The $\mathcal{M} \in \mathbb{R}^{N\times L\times L}$ is the connection mask:
\begin{equation}
\begin{aligned}
& \mathcal{M}_{i,j} = 1~~if~~\mathcal{C}_{i,j} >= 0,\\
& \mathcal{M}_{i,j} = 0~~if~~\mathcal{C}_{i,j} < 0,\\
\end{aligned}
\label{thred}
\end{equation}
\textbf{ii) Relevance.} Due to the heterogeneity between videos and images, simply using global embeddings for cross-modal relevance learning is insufficient. So we continue to employ fine-grained matching based on multiple nodes to narrow the cross-domain differences. 

For the video-to-image associations, we first extract the column vectors of $\mathcal{H}^i_{V2I}$ and then simultaneously compute the mean value and remap the top-K matching scores:
\begin{equation}
\begin{aligned}
 \mathcal{A}_i &= \operatorname{Mean}(\mathcal{H}^i_{V2I}) + \operatorname{\mathcal{F}_{MLP}}(\operatorname{topK}(\mathcal{H}^i_{V2I}, K)),\\
\end{aligned}
\label{rank_m}
\end{equation}
where $i \in [0,L)$ is the column index and $K=L$. Since the relevance learning for different samples within a batch is independent, we use an additional batch linear layer to encourage interactions between different samples and further expand the matching receptive field.
\begin{equation}
\begin{aligned}
 \mathcal{R}_i &= \operatorname{\mathcal{F}_{BLL}}( \mathcal{A}_i),\\
\end{aligned}
\label{rank_m}
\end{equation}
where the $\operatorname{\mathcal{F}_{BLL}}(\cdot)$ includes an ReLU activation layer and two linear layers. Then we obtain the relevance matrix $\mathcal{R}$ for all pairs of video-to-image instances. 

\begin{figure}[t]
    \centering
    \includegraphics[width=86mm]{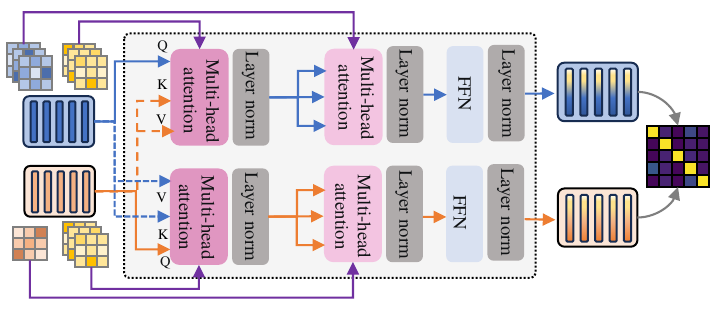}
    \vspace{-7mm} 
    \caption{Details of the cross-domain interaction module.}
    \label{CDI_block}
\vspace{-5.3mm} 
\end{figure}

Therefore, the cross-domain graph can be constructed from the connection matrix $\mathcal{C}$ and relevance matrix $\mathcal{R}$ as:
\begin{equation}
 \boldsymbol G_{V2I} = \mathcal{R}\otimes \mathcal{C}.\\
\end{equation}
\noindent\textbf{Cross-domain Interaction.} As shown in Fig.~\ref{CDI_block}, we design two attention branches to enhance the visual representation. The cross-domain connections in GCI represent the interaction of intra-domain and inter-domain features. We use the constructed graph $\boldsymbol G_{I2V}$ and $\boldsymbol G_{V2I}$ to replace the mask in the MHA function to guide the model attention to local areas with high relevance as:
\begin{equation}
\begin{aligned}
& I_{m}=\operatorname{\mathcal{F}_{LN}}(\operatorname{\mathcal{F}_{MHA}}({{\mathcal {Q}}_I}, {{\mathcal {K}}_V}, {\mathcal {V}}_V~|~ \boldsymbol G_{I2V})), \\
& V_{m}=\operatorname{\mathcal{F}_{LN}}((\operatorname{\mathcal{F}_{MHA}}({{\mathcal {Q}}_V}, {{\mathcal {K}}_I}, {\mathcal {V}}_I~|~ \boldsymbol G_{V2I})), 
\end{aligned}
\end{equation}
where $\operatorname{\mathcal{F}_{LN}}(\cdot)$ is the layer norm function. 
Then we take the graph-enhanced feature $I_{m}$ and $V_{m}$ through the MHA module under the guidance of the intra-modal graphs $\boldsymbol G_{I2I}$ and $\boldsymbol G_{V2V}$ to obtain the graph-enhanced visual representation $I_{g}$ and $V_{g}$, and $\operatorname{\mathcal{F}_{FFN}}(\cdot)$ is the function of a feed-forward network.
\begin{equation}
\begin{aligned}
& I_{g}=\operatorname{\mathcal{F}_{FFN}}(\operatorname{\mathcal{F}_{MHA}}({\mathcal {Q}_{I_{m}}}, {\mathcal {K}_{I_{m}}}, {\mathcal {V}_{I_{m}}} ~|~ \boldsymbol G_{I2I})), \\
& V_{g}=\operatorname{\mathcal{F}_{FFN}}(\operatorname{\mathcal{F}_{MHA}}({\mathcal {Q}_{V_{m}}}, {\mathcal {K}_{V_{m}}}, {\mathcal {V}_{V_{m}}} ~|~ \boldsymbol G_{V2V})).
\end{aligned}
\end{equation}

\textbf{Graph Loss.}
We first use the triplet loss $\mathcal{T}(\cdot)$ to compute the cosine similarity of features augmented by spatiotemporal graphs,  jointly optimizing the parameters of the image, video, and video-image graphs. In addition, the cross loss between the enhanced embedding and the original embedding is additionally calculated to narrow the feature domain difference, and the graph loss is:
\begin{equation}
\begin{aligned}
\mathcal{L}_{gr} = \mathcal{T}(V_{g}, I_{g}) + \mathcal{T}(V_{visual}, I_{g}) + \mathcal{T}(V_{g}, I_{visual}).
\end{aligned}
\end{equation}
We use the KL-divergence loss~\cite{van2014renyi} function $\mathcal{K}(\cdot)$ to minimize the difference between the distribution of calibrated cross-domain correlations and the global embedding.
\vspace{-1mm}
\begin{equation}
\mathcal{L}_{kl} = \mathcal{K}({\boldsymbol G}_{V2I}, \mathcal{H}_{V2I}) + \mathcal{K}({\boldsymbol G}_{I2V}, \mathcal{H}_{I2V}).
\vspace{-0.5mm}
\end{equation}
And the total loss in GCI is defined as $\mathcal{L}_{g} = \mathcal{L}_{gr} + \mathcal{L}_{kl}$.
\subsection{Selective Multi-modal Fusion}
To improve the discrimination of semantically related and visually similar products, we further perform the hard negative mining operation using the multi-modal features. First, we compute a similarity matrix with visual and textual representations in the GRA module to select hard examples as follows:
\begin{equation}
M_{sim}=V_{visual}\cdot{I^{\mathrm{T}}_{visual}} +\alpha V_{text} \cdot{I^{\mathrm{T}}_{text}},
\label{eq17}
\end{equation}
then we select top $K$ samples ranked in $M_{sim}$ as the hard samples for further matching. The $K$ is set to 4. The visual and textual features of these samples are selected as:
\begin{equation}
\begin{aligned}
 ind &= \operatorname{topK}(M_{sim}, K),\\
 \hat{V}_{visual}, \hat{I}_{visual} &= V_{visual}[ind], I_{visual}[ind],\\
 \hat{V}_{text}, \hat{I}_{text} &= V_{text}[ind], I_{text}[ind]. 
\label{rank_k}
\end{aligned}
\end{equation}

Besides, we fuse the textual feature and visual representations of selected samples using a fusion layer $\operatorname{\mathcal{F}_{fusion}}(\cdot)$. The multi-modal features of video and image are learned.
\begin{equation}
\begin{aligned}
& \hat{V} = \operatorname{\mathcal{F}_{fusion}}(\hat{V}_{visual}, \hat{V}_{text}),
& \hat{I}= \operatorname{\mathcal{F}_{fusion}}(\hat{I}_{visual}, \hat{I}_{text}). 
\end{aligned}
\end{equation}
The fused features $\hat{V}$ and $\hat{I}$ will pass through a perceptual module consisting of a self-attention layer and a cross-attention layer $\mathcal{F}_{cross}$ to obtain cross-domain features:
\begin{equation}
V_{cross} = \operatorname{\mathcal{F}_{cross}}(\mathcal {Q}_{\hat{I}}, {\mathcal {K}_{\hat{V}}}, {\mathcal {V}_{\hat{V}}}),
\end{equation}
Then we use the symmetric cross-entropy loss $\mathcal{E(\cdot)}$~\cite{fang2021clip2video} to calculate the matching loss in hard negative mining.
\begin{equation}
\mathcal{L}_{m} = \mathcal{E}(\operatorname{AvgPool}(V_{cross})).
\end{equation}
\subsection{Optimization Target} 
To learn a comprehensive representation from the video-image-text alignment space, the total loss is optimized as:
\begin{equation}
\begin{split}
\mathcal{L}_{total} = \mathcal{L}_{s}+ \beta_1 \mathcal{L}_{g}+\beta_2 \mathcal{L}_{m}.
\end{split}
\label{loss}
\end{equation}
Mentioned that the model only uses the independent embedding in GRA to calculate the global similarity in the inference stage, having high matching accuracy and efficiency.
\subsection{Model Inference} 
During the inference stage, only the GRA module is utilized to independently encode the global visual and textual representations of the input video and the product gallery images. The matching scores $\mathcal{S}$ are calculated as a weighted similarity of the embeddings from the visual and text domain. 
\begin{equation}
\mathcal{S} = V_{visual}I_{visual}^T+ \lambda V_{text}I_{text}^T.
\label{output}
\end{equation}

\section{Experiments}
\subsection{Experimental Setup}
\noindent\textbf{Datasets.} We conduct experiments on the publicly available dataset LPR4M~\cite{yang2023crossview} and MovingFashion (MF)~\cite{godi2022movingfashion}. The LPR4M dataset is the largest publicly available multi-modal dataset that covers 34 categories and comprises three modalities (image, video, and text). It includes 3,955,181 pairs of video-images for training and 20,079 pairs for testing, encompassing a diverse range of scenes. The MF dataset contains only one category of fashion clothing with two modalities (image, video), comprising 15,045 pairs in the training set and 1,341 pairs in the test set. These well-stocked commerce datasets facilitate a fair evaluation of our method in real-world scenarios. More dataset analyses are available in the supplements.

\noindent\textbf{Implementation Details.} We initialize image and video encoders using the pretrained ViT-B/32 model from CLIP~\cite{luo2022clip4clip} with a feature dimension of 512. The text encoder is initialized by the pretrained  RoBERTa-wwm-Base model from ChineseCLIP~\cite{yang2022chinese}. For video preprocessing, we extract 10 evenly spaced frames from each clip. Each frame in videos has a probability of 0.5 to be randomly masked for data augmentation, and the masking percentage ranges from 0 to 0.9. We set the weight factor $\lambda=0.5$ in Eq.~\ref{lamda} and Eq.~\ref{output}. The margin in triplet loss is set to 0.2 for proper discrimination. In Eq.~\ref{eq8}, top-K scores serve as the relevance threshold, set to $L/2$ to balance performance and complexity. In Eq.~\ref{eq17}, higher $K$ implies more memory (M) costs, so we set $K$=4. The weight factors are $\beta_1=0.7$ and $\beta_2=0.3$ in Eq.~\ref{loss}. In model optimization, we use the Adam~\cite{kingma2014adam} optimizer with a batch size of 256. The learning rate is $3 \times 10^{-4}$, which decays following the cosine schedule~\cite{loshchilov2016sgdr}. Our models are trained on 8 NVIDIA Tesla V100 GPUs. Following the standard retrieval task~\cite{zhang2018cross, mithun2018learning}, recall at rank K (R@K) is adopted as the metric to evaluate the performance quantitatively.
\begin{table}[t]
\vspace{-1mm}
\caption{Performance comparison on the LPR4M dataset.}
\vspace{-3mm}
\resizebox{0.47\textwidth}{!}{
\centering
\begin{tabular}{l|c|c|c|c}
    \hline
Methods   & R@1 & R@5 & R@10 & R@mean
    \\
    \hline   \hline
FashionNet~\cite{liu2016deepfashion} &13.4 & 33.8 & 50.4 & 32.5\\
AsymNet~\cite{cheng2017video2shop}  &22.0 & 46.7& 63.8  & 44.2\\
SEAM~\cite{godi2022movingfashion}  &23.3 & 49.5 & 61.4 & 44.7\\
TimeSFormer~\cite{bertasius2021space}  &28.6 & 56.8 & 69.0& 51.5\\
NVAN~\cite{liu2019spatially}  &21.4 & 45.2 & 62.7& 43.1\\
SwinB~\cite{liu2022video}  &29.1 & 60.1 & 73.9& 54.4\\
RICE~\cite{yang2023crossview}  &33.0 & 65.5 & 77.3 & 58.6\\
\hline \hline 
{SGMN (w/o TE)}&{38.7} & {66.5} & {76.2} & 60.5\\
\bf{SGMN (Ours)}&\bf{43.4} & \bf{68.9} & \bf{79.2} & \bf{63.8}\\
\hline   \hline
  \end{tabular}
}
\label{tab:lpr4m}
\vspace{-3mm}
\end{table}

\begin{table}[t]
\caption{Performance comparison on the MF dataset.}
\vspace{-3mm}
\resizebox{0.48\textwidth}{!}{
\centering
\Large
\begin{tabular}{l|c|c|c|c}
    \hline
Methods   & R@1 & R@5 & R@10& R@mean
    \\
    \hline   \hline
NVAN~\cite{liu2019spatially} &38.0 & 62.0 & 70.0 & 56.7 \\
MGH~\cite{yan2020learning}  &40.0 & 59.0& 66.0 & 55.0 \\
AsymNet~\cite{cheng2017video2shop}  &42.0 &73.0 & 86.0 & 67.0\\
SEAM~\cite{godi2022movingfashion}  &49.0 & 80.0 & 89.0 & 72.7\\
RICE~\cite{yang2023crossview}  &76.1 & 89.7 & 92.6 & 86.1 \\
\hline \hline
{SGMN (w/o Finetune)}&{43.1} & {63.5} & {70.7} & 59.1\\
\bf{SGMN (Ours)}&\bf{77.8} & \bf{90.3} & \bf{92.7} & \bf{86.9}\\
\hline   \hline
  \end{tabular}
}
\label{tab:mf}
\vspace{-5mm}
\end{table}

\begin{figure}[t]
\flushleft
\vspace{-0.25cm} 
\subfigbottomskip=-5pt 
\subfigcapskip=-5pt 
\subfigure[LPR4M dataset]{
\label{smf_a}
\centering
\includegraphics[width=82mm]{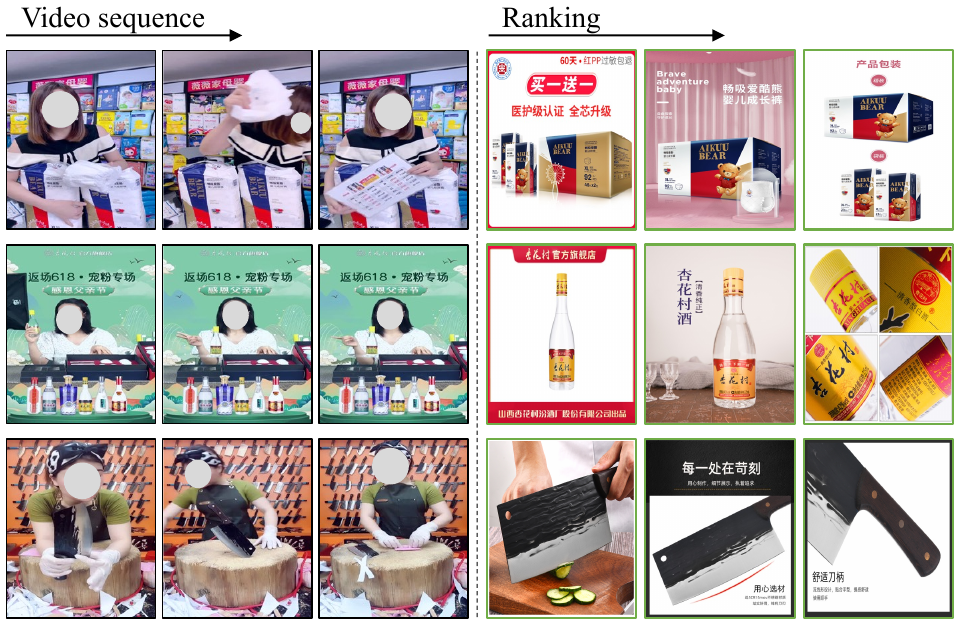}}
\subfigure[MF dataset]{
\label{smf_b}
\centering
\includegraphics[width=82mm]{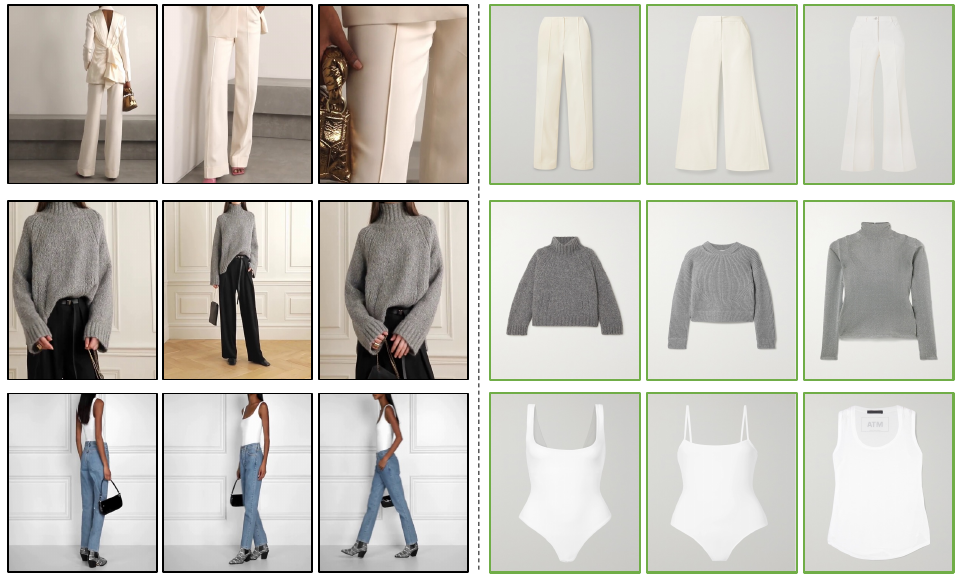}}
\vspace{-2.5mm}
\caption{Ranking results of representative products in LPR.}
\vspace{-3.5mm}
\label{fig:ranking}
\end{figure}
\begin{table}[t]
\vspace{-0.5mm}
\caption{Comparison of inference time of different methods.}
\vspace{-3.5mm}
\Large
	\resizebox{\columnwidth}{!}{
		\begin{tabular}{c|cccccc }
			\hline
    		 Method & AsymNet & SEAM &TimeSFormer  & RICE & SGMN\\
			\hline \hline 
			 Time (ms) & 910.4 & 622.6 & 163.2  & 24.0 & 24.3\\ 
              Params (M) & 295 &  52 &  185 & 158  & 174 \\ 
              R@1 (\%) & 22.0 & 23.3 &  28.6  &33.0&\textbf{43.4} \\ 
			\hline \hline 
		\end{tabular}
	}
	\label{tab:inference}
\vspace{-5mm}
\end{table}
\begin{table*}[t]
\caption{The R@1 performance of various methods is evaluated on the LPR4M dataset across subsets categorized by video variations, including product scale, visible duration, and number of products. The best performance for each subset is highlighted.}
\vspace{-4mm}
\begin{center}
\begin{tabular} {l|c|ccc|ccc|ccc}
\hline
\multirow{2}{*}{Methods} & \multirow{2}{*}{overall} & \multicolumn{3}{c|}{scale} & \multicolumn{3}{c|}{visible duration} & \multicolumn{3}{c}{number of product} \\
&  & small  &  medium  & large  &  short  &  medium  & long 
&  abundant  &  medium  & few  \\ 
\hline
    RICE$_{patch}$ & 31.2& 28.9  & {37.0} & 32.7  & 28.1 & 32.9 & {39.6}  & 21.0 & 34.6 & {31.5} \\
    RICE$_{box}$   & 33.0 & 32.7  & {39.0} & 33.8  & 29.6  & 34.8 & {42.0} & 17.6  & 31.9 & {33.4} \\
    \hline
    \textbf{SGMN (Ours)}  & \textbf{43.4} & \textbf{38.3}  & \textbf{43.0} & \textbf{54.9}  & \textbf{36.6}  & \textbf{43.6} & \textbf{54.5} & \textbf{24.7}  & \textbf{32.3} & \textbf{42.9} \\
\hline 
\end{tabular}
\end{center}
\label{tab:robustness}
\vspace{-3.5mm}
\end{table*}
\subsection{Comparison with Other Methods}
We quantitatively compare the proposed SGMN with the existing methods on the LPR4M dataset and MF dataset, and the results are shown in Tab.~\ref{tab:lpr4m} and Tab.~\ref{tab:mf}. On the large-scale dataset LPR4M, the retrieval performance of our method outperforms the state-of-the-art RICE~\cite{yang2023crossview} by 10.4\% in R@1, 3.4\% in R@5, and 1.9\% in R@10. Existing methods exhibit limited performance on LPR4M with rich categories and diverse scenes. We also verified the performance of SGMN without using textual features (w/o TE). Our SGMN can still achieve a performance gain of 5.7\% on R@1 without the textual guidance to the salient regions (38.7\% vs 33.0\%). 

Since the MF dataset does not have text modality, we do not fuse the textual information in training. Efficient global cross-domain alignment still allows SGMN to maintain the best performance with significant gains. Compared with the official method of MF dataset SEAM, our method improves R@1 by 28.8\% in R@1. Although the MF testset contains only 1,342 videos, our approach still gains a 1.7\% advantage over the highest RICE method in the crucial R@1 metric, signifying our superiority in extracting fine-grained discriminative features. We also verified the model trained on LPR4M without finetuning on MF, and the results show that even if the model has not seen the MF data, it can still surpass most trained models, demonstrating robust zero-shot fitting and generalization capability. To summarize, our method benefits from cross-domain connections and fine-grained matching, showing greater superiority on both simple and diverse datasets, particularly in the crucial R@1 metric.
\begin{figure}
    \centering
    \includegraphics[width=84mm]{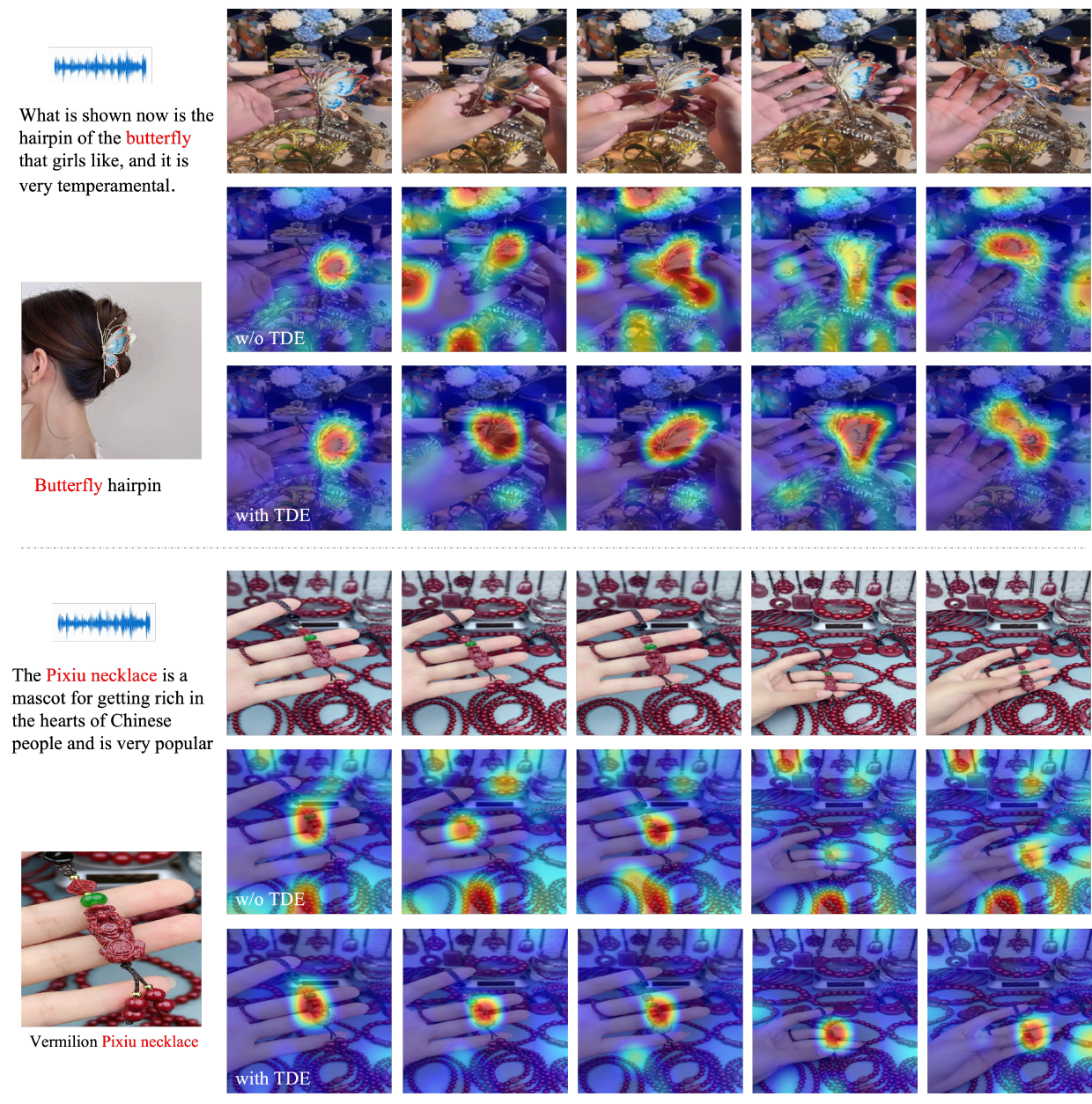}
    \vspace{-4mm} 
    \caption{Attention visualization with and without (w/o) TE modules. Benefiting from the embedding and guidance of text information, the model tends to focus on keyword-related products in the cluttered background. The keywords `butterfly’ and `Pixiu necklace' that co-exist in the video text and product items correct the model attention efficiently.}
    \label{butterfly}
\vspace{-4.5mm} 
\end{figure}
\vspace{-2mm}
\subsection{Qualitative Results}
\noindent\textbf{Retrieval Results.} We randomly select several video sequences from MF dataset and LPR4M dataset, and the top-3 shop images are shown in Fig.~\ref{fig:ranking}. It can be seen that even if the live scene is filled with numerous products of the same type, our SGMN can still find the best matched product. Besids, our model performs excellently in distinguishing highly similar clothes in the MF dataset. Even though the intended products encounter appearance distortion due to occlusion, motion, scaling, or illumination variations, our method can still accurately retrieve the correct products.

\noindent\textbf{Inference Time.} We show the model inference time in Tab.~\ref{tab:inference} to verify the efficiency of our one-stage strategy. Experiments for different methods were conducted following their open-source setting and replicated on the Tesla T4 GPU. Compared to two-stage methods like AsymNet and SEAM, which require additional object detection, our SGMN achieves over 30 times faster speed. Even with the TE module, our SGMN demonstrates nearly identical runtime to RICE using the same CLIP model. Benefiting from globally independent encoding and parameter sharing, our SGMN balances well between model complexity and performance.

\noindent\textbf{Generalization and Robustness.}  We carried out experiments on different video variations of the LPR4M dataset and the results are in Tab.~\ref{tab:robustness}. RICE$_{patch}$ is a one-stage algorithm, while RICE$_{box}$  is a two-stage algorithm manually incorporating detected boxes. Data is divided into small, medium, and large subsets with different product scales, visible durations, and product numbers. Our SGMN consistently outperforms RICE by a significant margin across various subsets, demonstrating robustness in real-world scenarios.
\begin{table}[t]
\footnotesize
	\centering
    \caption{Hyperparameters analysis of loss function.}
    \vspace{-3.5mm}
	\resizebox{\columnwidth}{!}{
		\begin{tabular}{c|cccccccc}
			\hline \hline
    		$\beta_1$ & 0 & 0.2 & 0.3 & 0.5 & 0.7 & 0.8 & 1\\
            $\beta_2$ & 1 & 0.8 & 0.7 & 0.5 & 0.3 & 0.2 & 0\\
			\hline 
			R@1 & 39.1 & 41.8 & 40.7 & 41.5 & \textbf{43.4}  & 39.4 & 40.9\\ 
			\hline \hline
		\end{tabular}
	}
	\label{tab:loss}
\vspace{-4mm}
\end{table}
\begin{table}
\small
	\centering
 \caption{Performance of combining different relation graphs, including intra- and inter-domain graphs of video and image.}
  \vspace{-3.5mm}
	\resizebox{\columnwidth}{!}{
		\begin{tabular}{c|ccc|ccc }
			\hline \hline
    		$\#$ & $\boldsymbol G_{I2I}$ & $\boldsymbol G_{V2V}$ & $\boldsymbol G_{I2V + V2I}$  & R@1 &R@5 &R@10\\
			\hline 
			A & & &  &38.6 & 65.7 & 76.5\\ 
			B & $\checkmark$&  & & 38.8 & 65.9& 77.0 \\ 
            C & & $\checkmark$ & &39.7 & 66.7& 77.3\\ 
            D & & & $\checkmark$& 40.1& 67.0 & 77.8\\ 
			E &$\checkmark$ & $\checkmark$&  & 39.9& 67.2 & 77.6\\ 
		    H &$\checkmark$ & $\checkmark$& $\checkmark$ & \bf{43.4} & \bf{68.9} & \bf{79.2}\\
			\hline \hline
		\end{tabular}
	}
	\label{tab:graph}
 \vspace{-4mm}
\end{table}
\begin{figure*}[t]
\flushleft
\subfigtopskip=2pt 
\subfigbottomskip=2pt 
\subfigcapskip=-3pt 
\subfigure[Video sequences and product images of hard examples.]{
\label{smf_a}
\includegraphics[width=77mm]{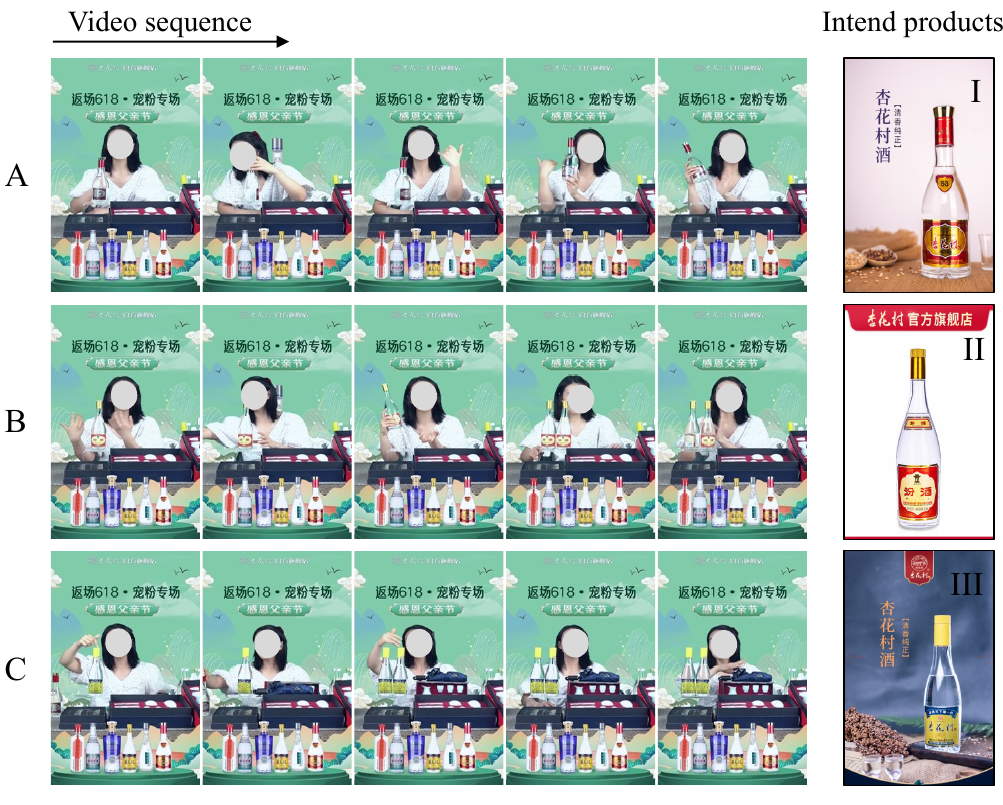}}
\subfigure[Global similarity]{
\label{smf_b}
\centering
\includegraphics[width=31mm]{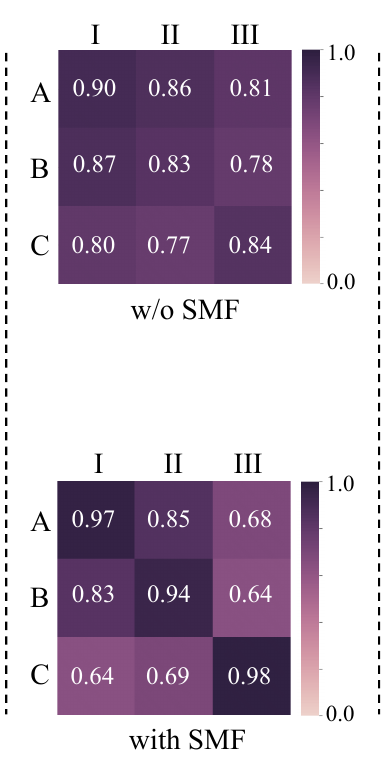}}
\subfigure[Attention visualization for retrieved videos]{
\label{smf_c}
\centering
\includegraphics[width=61mm]{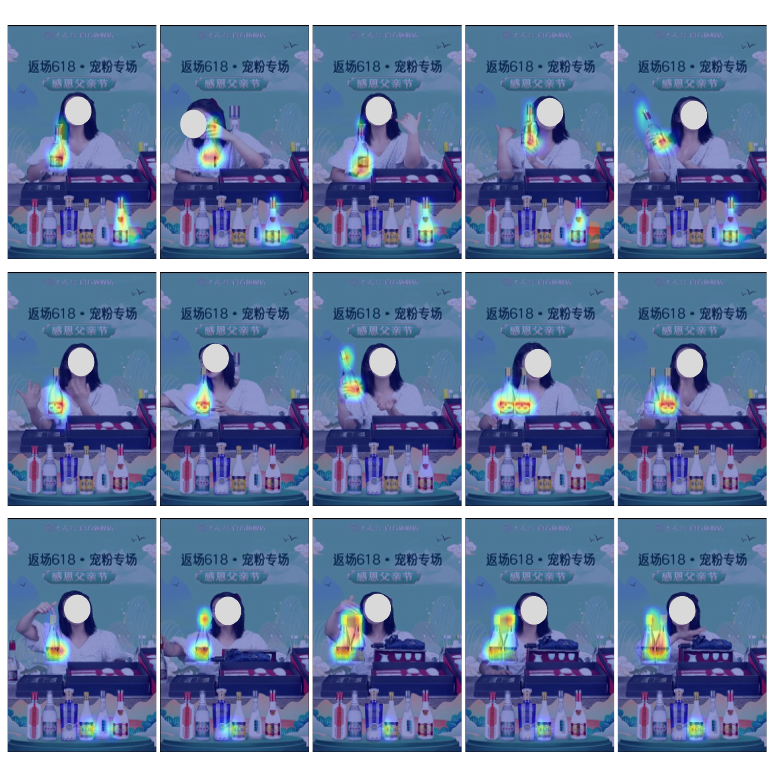}}
\vspace{-4mm}
\caption{Quantitative and qualitative comparison of our proposed SMF module in identifying multiple highly similar products. (a). Three similar-looking liquor video sequences (A, B, and C) and corresponding intended product images (I, II, and III).  (b). Quantification of the global embedding similarity between the three videos and the corresponding product images using the model with or without the SMF module. (c). Attention visualization of our SGMN on encoded features for three video sequences.}
\vspace{-2.5mm}
\label{smf_abc}
\end{figure*}
\vspace{-6mm}
\subsection{Ablation Study}
\noindent\textbf{Text Attention Analysis.} 
Embedding text from video ASR and product items can guide the model to focus on the intended item in the cluttered background at a low cost. Therefore, we visualized the attention of the video encoder in GRA equipped with or without (w/o) the TE module. As shown in Fig.~\ref{butterfly}, the method with TE is superior to that w/o TE in accurately retrieving the intended product. For example, in the live clip of selling hairpins, multiple similar hairpins appear in view simultaneously, interfering with the identification of intended products. In this case, the keyword `butterfly' mentioned by the live salesperson played a key role, significantly increasing the probability of matching products that contain both `butterfly' and `hairpin' in the title. Similarly, there are a large number of extremely similar products in the vermilion necklace category gallery, and their visual differences are almost negligible in the live domain, where the resolution is degraded. Fortunately, the assistance of text information solves this dilemma very well. The keywords `Pixiu' and `necklace' have endowed the product with specificity, freeing the network from the interference of redundant products irrelevant to the keywords. These results show that our SMGN equipped with additional textual representations alleviates the difficulty of product-heavy recognition.

\noindent\textbf{Loss Function Analysis.} As shown in the Tab.~\ref{tab:loss}, we analyze the weights of different components in the loss function. It can be observed that not using graph loss or hard example loss is unwise. A higher weight for the graph loss brings more performance gain, but it needs to be balanced with the SMF module. The frame-level temporal correlation complements high similarity discrimination, so we set $\beta_1$ and $\beta_2$ in Eq.~\ref{loss} as 0.7 and 0.3 respectively.

\noindent\textbf{Different Spatiotemporal Graphs.} To analyze the performance of spatiotemporal graph learning in enhancing cross-domain interaction, we verified the performance using different relation graphs, and the results are shown in Tab.~\ref{tab:graph}. The baseline uses the self-attention mechanism to interact with the initial features of videos and images. It can be seen that using graph learning to enhance intra-domain correlation is helpful because the methods using $\boldsymbol G_{I2I}$ or $\boldsymbol G_{V2V}$ graphs both have performance gains. Besides, constructing $\boldsymbol G_{I2V}$ and $\boldsymbol G_{V2I}$ graphs with cross-domain interactions has a higher gain in R@1 (40.1\% vs 38.6\%), indicating that the spatiotemporal aligned domain is more conducive to learning efficient representations. Since products in the live domain often have appearance deformations, capturing enough fine-grained and long-range features to distinguish video-to-image pairs accurately is critical. The jointly modelled temporal consistency and spatial correlation enhance the global embedding of images and videos, achieving a 4.8\% accuracy gain in R@1 (43.4\% vs 38.6\%).

\noindent\textbf{Identification of Similar Products.} 
In livestreaming e-commerce, similar products often appear in the same field of view to facilitate sales, further exacerbating recognition accuracy. To illustrate that our proposed SMF module strengthens the model in distinguishing hard negative samples, we selected several video sequences with top scores in the `Liquor' category retrieval for quantitative and qualitative comparisons. As shown in Fig.~\ref{smf_a}, the product images of the three liquors have extremely slight visual differences, and their corresponding live video sequences are also easily confused. We quantify the similarity scores of the three videos (A, B, and C) and the corresponding product images (I, II, and III) with and without (w/o) the SMF module, and the results are shown in Fig.~\ref{smf_b}. It can be seen that the model without SMF has poor scoring discrimination for similar products, and the scores of the correct category and the wrong category are very close, which increases the probability of misclassification. On the contrary, the model that further performs hard example mining using SMF has higher discriminative power for similar products. Furthermore, we visualize the visual attention of our SGMN on three video sequences in Fig.~\ref{smf_c}. The mining of hard samples incorporating multi-modal features captures sufficient fine-grained features to recalibrate model attention on intended products accurately and efficiently.



\vspace{-6mm}
\section{Conclusion}
In this paper, we rethink the LPR task from a more macroscopic and practical point of view and propose a one-stage spatiotemporal graphing network to address the real-world dilemmas of LPR tasks. To the best of our knowledge, this is the first exploration of sequence-to-sequence graph learning in mitigating heterogeneity between videos and images. Existing methods are limited by their narrow focus on coarse-grained matches. But we progressively enhance the fine-grained discrimination by integrating \textit{modality-level} text embeddings, \textit{instance-level} similarity mining, and \textit{frame-level} graph learning. Despite potential distortions in appearance within the complex livestreaming domain, our method adeptly tracks spatial deformations, ensuring precise location of the intended product. Moreover, we have fully harnessed the accessible textual information from live ASR transcripts and product titles, freeing the network from interference over cluttered background products. The further hard negative mining in video-image-text alignment domain improves the ability of our model to distinguish highly similar products. Extensive quantitative experiments show that our method well satisfies the demand for both local fine-grained attention and global spatiotemporal awareness in real-world scenarios. 





\bibliographystyle{ACM-Reference-Format}
\bibliography{ref}

\clearpage
\appendix
\section{Appendix}



\noindent
In this section, we provide the following supplementary materials that are not included in our main paper due to space constraints:
\begin{itemize}
\item More dataset analysis.
\item Experimental details on MF dataset.
\item Principle of loss function.
\item More model analysis.
\item More visualization results.
\end{itemize}

\subsection{More Dataset Analysis}
Both MF and LPR4M datasets used in our paper are publicly available. We analyze the properties of the two datasets, including data modality, number of product categories, number of videos and images, and number of pairs of (video, image). The quantitative comparison results are shown in Tab.~\ref{tab:dataset}. MF is a dataset containing only one category of fashion clothing, with two modalities, video (V) and image (I). And LPR4M is the largest publicly available LPR dataset that covers 34 commonly used live e-commerce categories, with three modalities, video, image, and text (T). Our proposed SGMN can efficiently and comprehensively utilize multiple modal information while maintaining the perception of fine-grained features. The capture of long-range spatiotemporal correlation enables our model to identify intra-class differences in clothing data and accurately distinguish inter-class diversity among multiple classes, achieving SOTA performances on both datasets above. 
\begin{table}[hp!]
\vspace{-2mm}
\caption{Analysis of MF and LPR4M datasets.}
\vspace{-3mm}
\tiny
\centering
\resizebox{0.42\textwidth}{!}
{
\begin{tabular}{c|c|c|c}
\hline
 & Property & MF & LPR4M \\
\hline 
\multirow{2}{*}{Content}& Modal& V \& I & V \& I \& T \\
& Category& 1 & 34 \\
\hline
\multirow{3}{*}{Training}&Videos& 15,045 & 3,955,181 \\
&Images& 14,855 & 272,063 \\
&Pairs& 15,045 & 3,955,181 \\
\hline
\multirow{2}{*}{Testing}&Videos&  1,342 & 20,079 \\
&Images& 1,341 & 66,358 \\
\hline
\end{tabular}
}
\label{tab:dataset}
\vspace{-4mm}
\end{table}
\subsection{Experimental Details on MF Dataset}
Regarding the fine-tuning details on the MF dataset, we initialize the weights using the model pretrained on the LPR4M dataset. The initial learning rate is $1 \times 10^{-4}$. The model is trained for 20 epochs on 8 NVIDIA Tesla V100 GPUs within only 4 hours. Other settings are maintained the same as those used in training on LPR4M, as described in the Experiments section of the main paper. To facilitate thorough testing, we redeveloped the evaluation code, which will also be included in the code submission. Evaluation is conducted using the model from the last epoch. For both training and testing, we extract 10 evenly spaced frames from each video as input.



\vspace{-1.5mm}
\subsection{Principle of Loss Function}

\noindent\textbf{Triplet loss.} The principle of triplet loss is to optimize the distance between anchor examples and positive examples to be smaller than the distance between anchor examples and negative examples, which is calculated as:
\begin{equation}
\mathcal{X}(a, p, n)=\max ({d}(a, p)-{d}(a, n)+\operatorname{margin}, 0),
\end{equation}
where $a$ is an anchor, $p$ and $n$ are positive and negative samples, $d(\cdot)$ is distance metric function and $\operatorname{margin}$ is a constant greater than 0, and we set the $\operatorname{margin}$ as 0.2. 

In Eq.5 of main paper, for the obtained global video representations $V_{visual} = \{v_1, v_2, ... ,v_N\}  \in \mathbb{R}^{N\times D}$ and image representations $I_{visual} = \{i_1,i_2, ... ,i_N\} \in \mathbb{R}^{N\times D}$, we define each sample in $V_{visual}$ and $I_{visual}$ as $v_j \in \mathbb{R}^{1\times D}$ and $i_k\in \mathbb{R}^{1\times D}$. Then we compute the cosine similarity matrix between $V_{visual}$ and $I_{visual}$ as follow:
\begin{equation}
M_{sim} = V_{visual}I_{visual}^T,
\end{equation}
where each element in $M_{sim}$ is $m_{jk} = v_j i_k, j,k \in [0,N]$.
Then we determine the position of the positive and negative samples in the $M_{sim} \in \mathbb{R}^{N\times N}$. When $j=k$, $(v_j,i_k)$ is a pair of positive samples, and the rest are negative samples:
\begin{equation}
\begin{aligned}
& (v_j,i_k)^{+} = 1~~if~~j=k,\\
& (v_j,i_k)^{-} = 0~~if~~j\neq k,\\
\end{aligned}
\label{thred}
\end{equation}
We optimize the distance between positive and negative samples so that positive samples are close to each other and negative samples are far away from each other:
\begin{equation}
\begin{aligned}
&{d}(\boldsymbol{v}, \boldsymbol{i}, \boldsymbol{i}^{-})=\max (\left[\operatorname{margin}-s(\boldsymbol{v}, \boldsymbol{i})+s\left(\boldsymbol{v}, \boldsymbol{i}^{-}\right)\right], 0)\\
&{d}(\boldsymbol{v}, \boldsymbol{i}, \boldsymbol{v}^{-})=max(\left[\operatorname{margin}-s(\boldsymbol{v}, \boldsymbol{i})+s\left(\boldsymbol{v}^{-}, \boldsymbol{i}\right)\right], 0),
\end{aligned}
\end{equation}
where $s(\boldsymbol{v}, \boldsymbol{i})$ is denoted as the cosine similarity. Thus the triplet loss between $V_{visual}$ and $I_{visual}$ is:
\begin{equation}
\begin{aligned}
\mathcal{T}(V_{visual}, I_{visual})& =\mathcal{X}(a,(\boldsymbol{v},\boldsymbol{i})^+, (\boldsymbol{v},\boldsymbol{i})^- ) \\
& ={d}(\boldsymbol{v}, \boldsymbol{i}, \boldsymbol{i}^{-})+ {d}(\boldsymbol{v}, \boldsymbol{i}, \boldsymbol{v}^{-}).
\end{aligned}
\end{equation}
Similarly, we calculate $\mathcal{T}(V_{text}, I_{text})$ as the same way.

\noindent\textbf{Symmetric Cross-entropy Loss}
In the SMF module, we calculate the symmetric cross-entropy loss based on the fusion feature $\mathcal{L}_{m} = \mathcal{E}(\operatorname{AvgPool}(V_{cross}))$ in Eq.21. Each training batch $\Omega$ consists of $N$ video-image pairs $(\hat{v}_m, \hat{i}_n)$, and they are discriminated in cross-entropy loss as follows:

\begin{equation}
\begin{gathered}
\mathcal{F}_{i 2 v}=-\frac{1}{N} \sum_{m \in \Omega} \log \frac{\exp(u_{mm})}{\sum_{n \in \Omega} \exp (u_{mn})}, \\
\mathcal{F}_{v 2 i}=-\frac{1}{N} \sum_{n \in \Omega} \log \frac{\exp (u_{nn})}{\sum_{m \in \Omega} \exp (u_{nm})))},
\end{gathered}
\end{equation}
where $u_{mn}$ is the element in $\hat{V}_{cross}$. Thus, the weighted sum of these two cross-entropy losses is the symmetric cross-entropy loss, corresponding to Eq.21 of the main paper:
\begin{equation}
\mathcal{E}(\operatorname{AvgPool}(V_{cross})) =\frac{1}{2}\left(\mathcal{F}_{i 2 v}+\mathcal{F}_{v 2 i}\right).
\end{equation}

\vspace{-2mm}
\subsection{More Model Analysis}
\noindent\textbf{Model Component Ablation.} 
We analyze the performance of each component in our method on the LPR4M dataset and present the quantitative results in Tab.~\ref{tab:block}. The baseline is a GRA module that independently encodes visual representations. It can be seen that each component we designed is effective. Among them, the TE introducing text attention and GCI learning spatiotemporal cross-domain interaction achieved higher performance gains, which are 4.5\% and 2.7\% in R@1, respectively. TMC enhances the intra-frame correlation of the global embedding of the video, and SMF further distinguishes similar samples, achieving R1 gains of 0.9\% and 2.5\%. Overall, the GCI module emphasizes frame-level fine-grained correlation over global coarse-grained alignment, while the SMF module achieves highly similar semantic discrimination at minimal cost. Our proposed solution aims to comprehensively optimize spatiotemporal consistent multi-modal representations, addressing the application dilemma of LPR step by step.

\noindent\textbf{Analysis of Weighting Factor.}
We analyze the weight factor $\lambda$ in the similarity loss function (Eq.5 of the main paper) to verify the contribution of textual information in the TE module. Specifically, we set the weight factor as $\lambda=\{0,0.3,0.5,0.7,0.9,1\}$ and show the results in Tab.~\ref{tab:loss}. It can be seen that the use of text domain information can obtain considerable performance gains. Since the automatic speech recognition (ASR) information in the livestreaming domain usually contains excessive verbal habits and unrelated clutter information, using more weight to the text embedding will degrade the performance.
\begin{table}[t]
\small
	\centering
 \caption{Model component ablation on the LPR4M dataset.}
  \vspace{-3.5mm}
	\resizebox{\columnwidth}{!}{
		\begin{tabular}{c|cccc|ccc }
			\hline
    		$\#$ & TE & TMC & GCI  & SMF & R@1 &R@5 &R@10\\
			\hline 
			A & & & & &32.8 & 58.7 & 72.7\\ 
			B & $\checkmark$& & & &37.3 & 63.8 & 74.4\\ 
			C &$\checkmark$ & $\checkmark$& & &38.2 & 65.2 & 76.1\\ 
		    D &$\checkmark$ & $\checkmark$& $\checkmark$& &40.9 & 67.3 & 77.8\\ 
			E &$\checkmark$ &$\checkmark$ &$\checkmark$ &$\checkmark$ & \bf{43.4} & \bf{68.9} & \bf{79.2}\\ 
			\hline
		\end{tabular}
	}
	\label{tab:block}
 \vspace{-2mm}
\end{table}

\begin{table}[t!]
\caption{Analysis of the weighted factor $\lambda$ in similarity loss.}
\vspace{-3mm}
\flushleft
\Small
\resizebox{0.46\textwidth}{!}{
\begin{tabular}{c|cccccc}
\hline 
$\lambda$ &0 & 0.3& 0.5& 0.7 & 0.9& 1.0\\
\hline
R@1 & 38.7  & 40.4 &  \textbf{43.4}& 41.2& 39.9 &  40.9 \\
R@5 & 66.5 & 67.8 &  \textbf{68.9} & 68.1 & 66.8  &  67.3 \\
R@10 & 76.2 & 77.9  &  \textbf{79.2} & 78.1& 77.1 &  77.6 \\
\hline
  \end{tabular}
}
\label{tab:loss}
\vspace{-2mm}
\end{table}
\begin{table}[t]
\centering
\caption{Performance of different feature fusion ways.}
\vspace{-3mm}
\tiny
\resizebox{0.49\textwidth}{!}{
\begin{tabular}{l|c|c|c}
\hline 
Fusion   & R@1 & R@5 & R@10\\
\hline
None & 39.7 & 66.7  &  76.2 \\
Addition & 38.7 &  65.8 &  75.0 \\
Concatenation & 40.3 & 68.3  &  78.0 \\
\hline
\bf{Cross-attention (Ours)}&\bf{43.4} & \bf{68.9} & \bf{79.2}\\
\hline 
  \end{tabular}
}
\label{tab:mmf}
\vspace{-2mm}
\end{table}

\noindent\textbf{Multi-modal Feature Fusion Mechanism.}
We analyze different ways of fusing features from visual and text modalities, including feature addition, feature concatenation, and cross-attention fusion, and present their results in Tab.~\ref{tab:mmf}. Due to the heterogeneity of visual and textual features, the practice of directly adding multi-modal features will make the global embedding misaligned, resulting in suboptimal performance. The way of feature aggregation using feature concatenation can make features from different modalities be projected into different spaces and optimized to obtain performance gains. However, this aggregation in the feature dimension cannot perform cross-domain interactions, so the benefits to performance are limited. Therefore, we design a cross-attention layer to introduce adaptively learned parameters to optimize the cross-modal fusion of text and visual features. Experimental results demonstrate that the cross-attention-guided feature fusion layer alleviates cross-domain heterogeneity and recalibrates the multi-modal information, achieving the highest performance gain.

\subsection{More Visualization Results}
To better demonstrate the practicability of our method in real-world application scenarios of livestreaming, we provide more visualization results of livestreaming domain.
\begin{figure}[t]
    \centering
    \includegraphics[width=85.5mm]{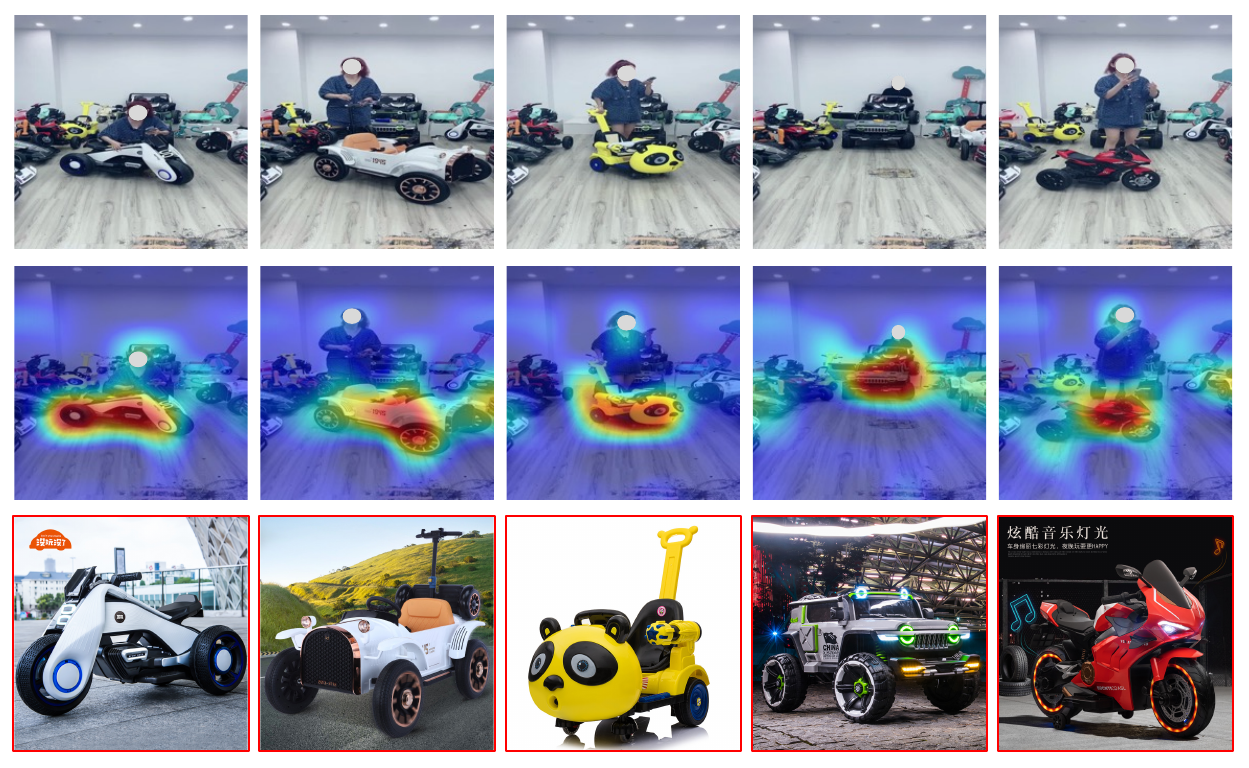}
    \vspace{-7mm} 
    \caption{A real-world livestreaming scenario in the LPR4M dataset that the salesperson puts all the products to be sold but explains them one by one. Images with red boundary denote the Top-1 ranking results from our SGMN.} 
    \label{sup_toy}
\vspace{-2mm}
\end{figure}
\begin{figure*}[t]
    \centering
    \includegraphics[width=0.94\linewidth]{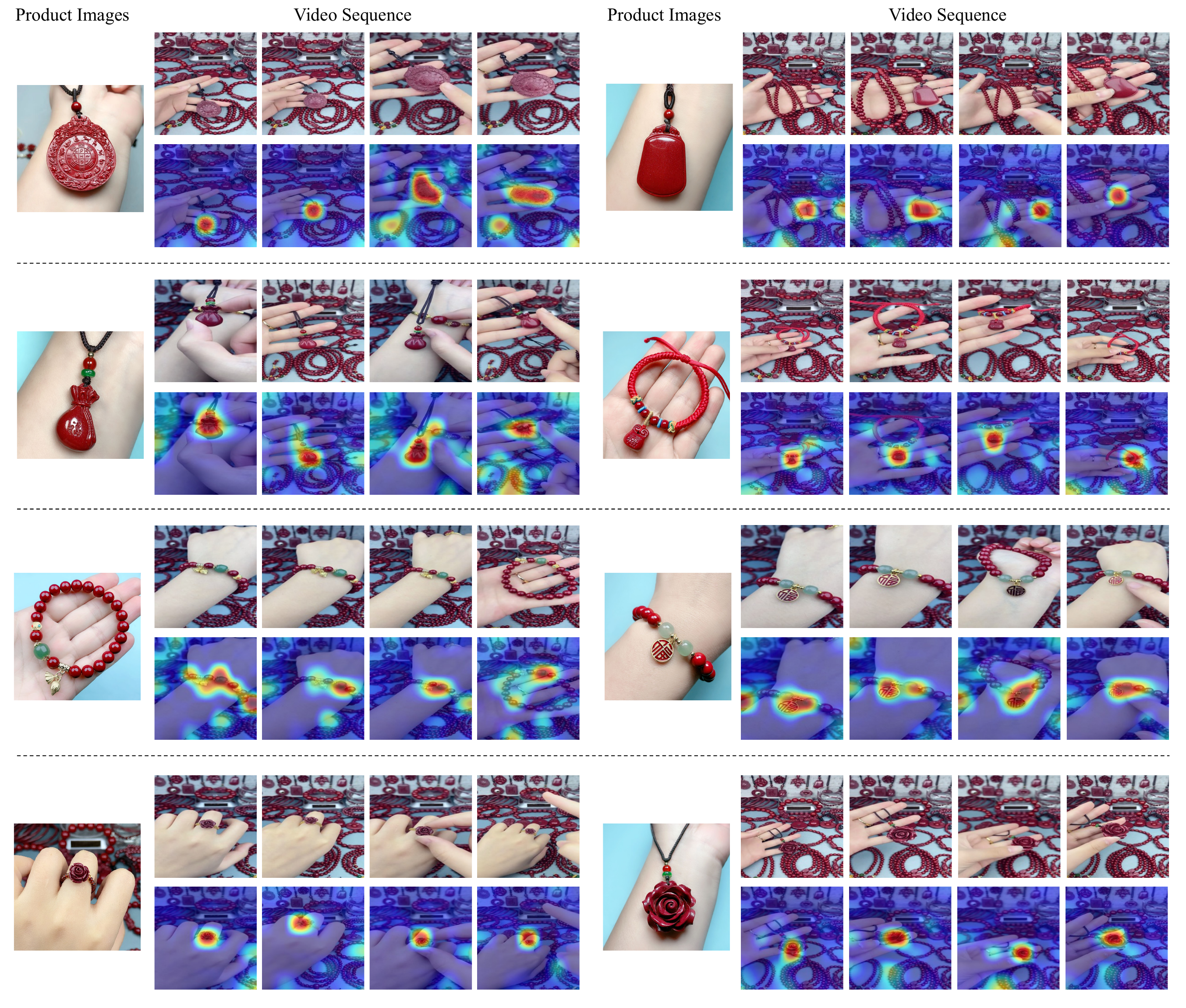}
    \vspace{-5mm} 
    \caption{Attentional visualization of highly similar products in the cluttered background scenario. Even if similar-looking Zisha ornaments appear simultaneously in the live domain, our model can still focus on the intended product in each frame.} 
    \label{sup_shouchuan}
\vspace{-3mm} 
\end{figure*}

\noindent\textbf{Products in cluttered background.} In the live clip shown in Fig.~\ref{sup_toy}, the salesperson showed a variety of toy cars at the same time, but only sold one product at a specific time. The dual guidance of text and visual features enables our method to accurately retrieve the expected product even if there are multiple similar products. As shown in Fig~\ref{sup_shouchuan}, we show the attention visualization of 8 similar-looking Zisha ornaments from the LPR4M dataset in the live domain. Specifically, we project the feature values of the last layer in the video encoder in Global Representation Alignment (GRA) module to [0,1], and draw the heatmaps for the corresponding video frames. The attention maps show which areas the model focuses on, and the significant regions are marked as highlighted, where red means significant, orange, green, and blue indicate that the importance gradually decreases in order. Our SGNM still has higher discriminative ability for intended products even if the background has multiple highly similar and easily confused products. 

\begin{figure*}[t]
    \centering
    \includegraphics[width=0.87\linewidth]{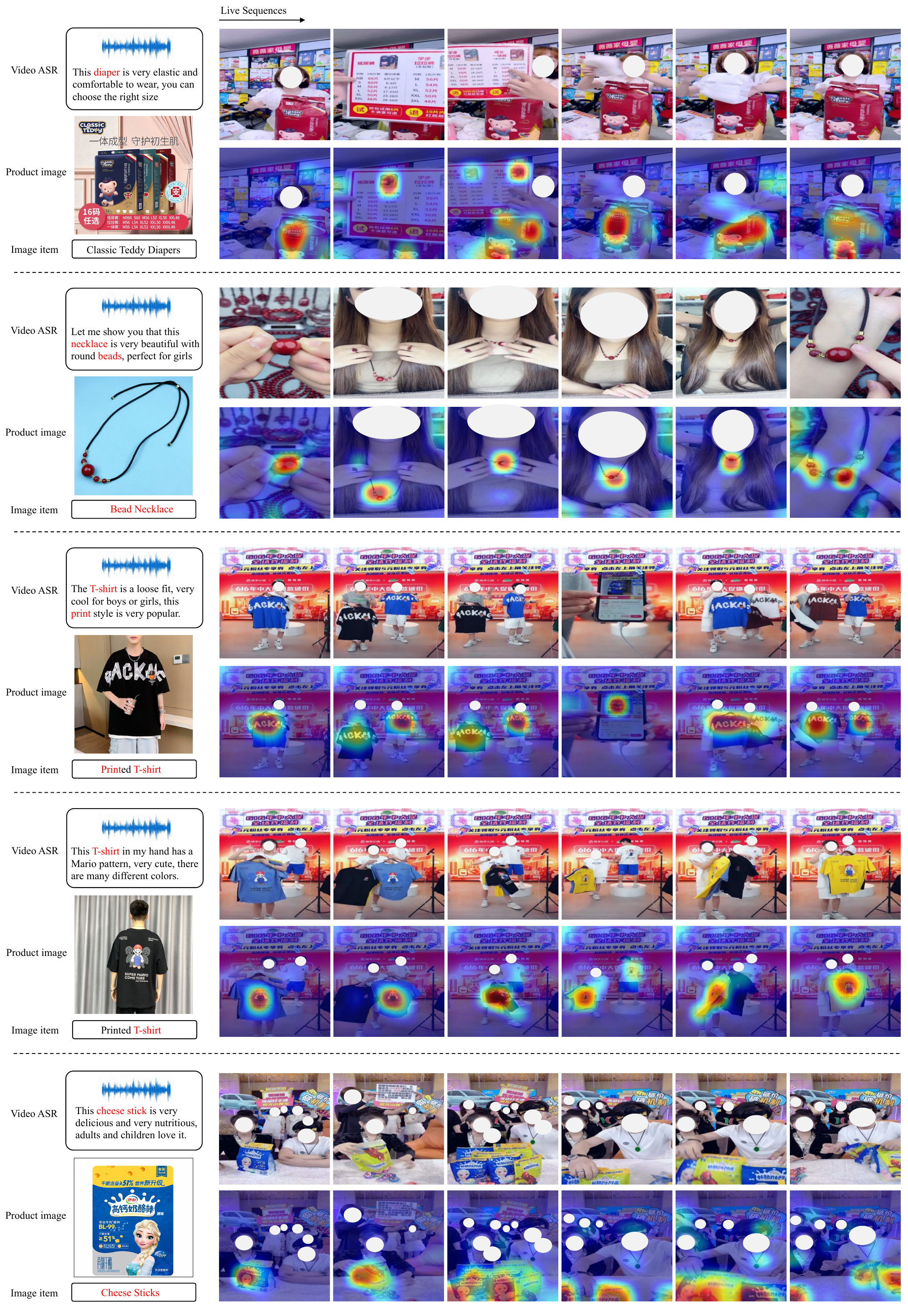}
    \vspace{-3mm} 
    \caption{Representative examples to demonstrate that our method consistently focuses on the correct region for retrieving intended products in the livestreaming domain, even with appearance variations and structural deformation.} 
    \label{sup_f1}
\vspace{-3mm} 
\end{figure*}

\noindent\textbf{Products with Appearance Variations.} The results in Fig.~\ref{sup_f1} show that our model can still accurately localize regions of intended products that encounter appearance distortion due to occlusion, motion, scaling, or illumination variations. The results in Fig.~\ref{sup_f1} show that our model can guide the model to focus on the intended products from the LPR4M dataset in the cluttered background and distinguish highly similar products with fine-grained features. 

\begin{figure*}[t!] 
    \centering 
    \subfigbottomskip=-1pt
    \subfigcapskip=-2pt 
    \subfigure[]{
        \label{sup_background}
        \centering
        \includegraphics[width=0.99\linewidth]{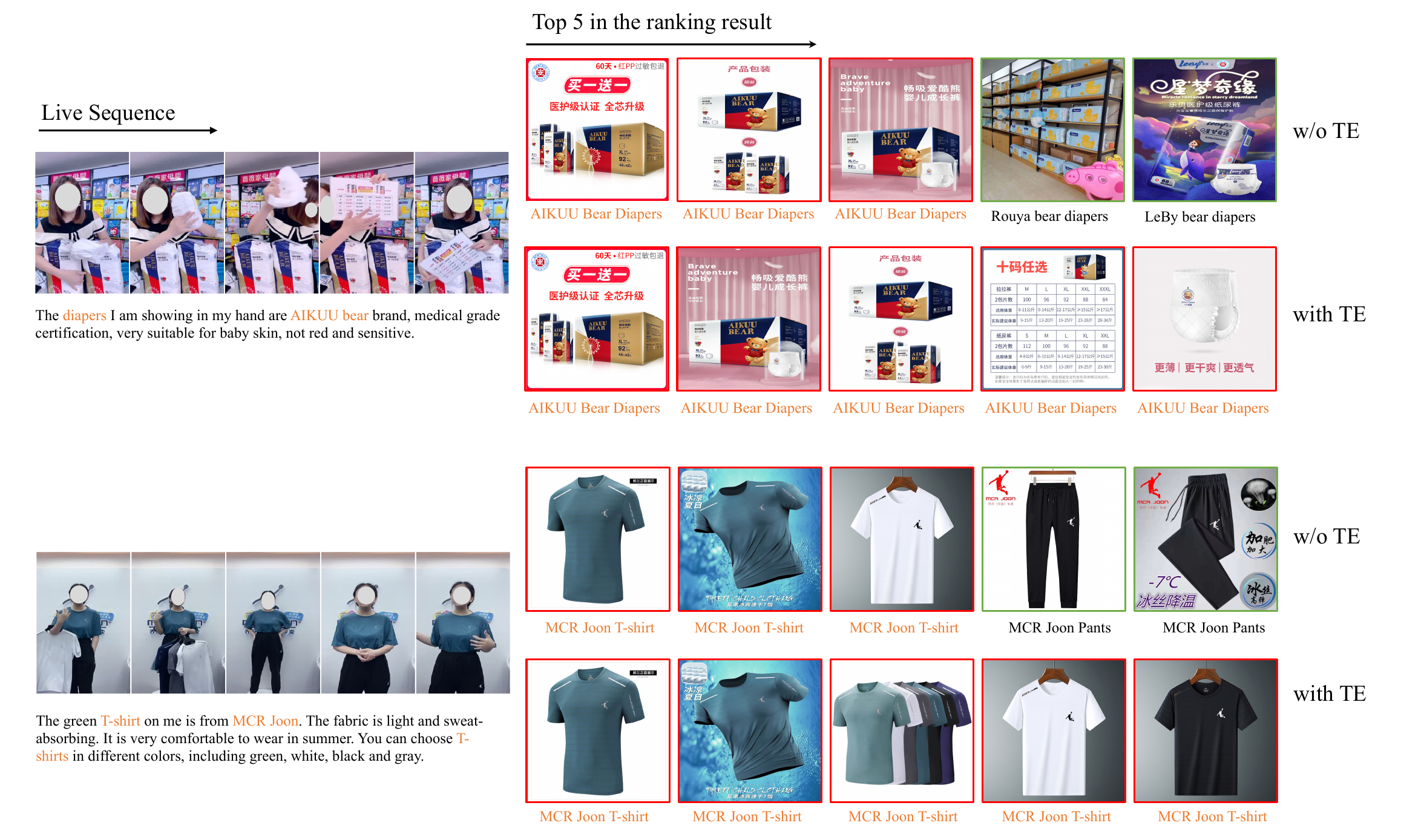}}
    \subfigure[]{
        \label{sup_similar}
        \centering
        \includegraphics[width=0.99\linewidth]{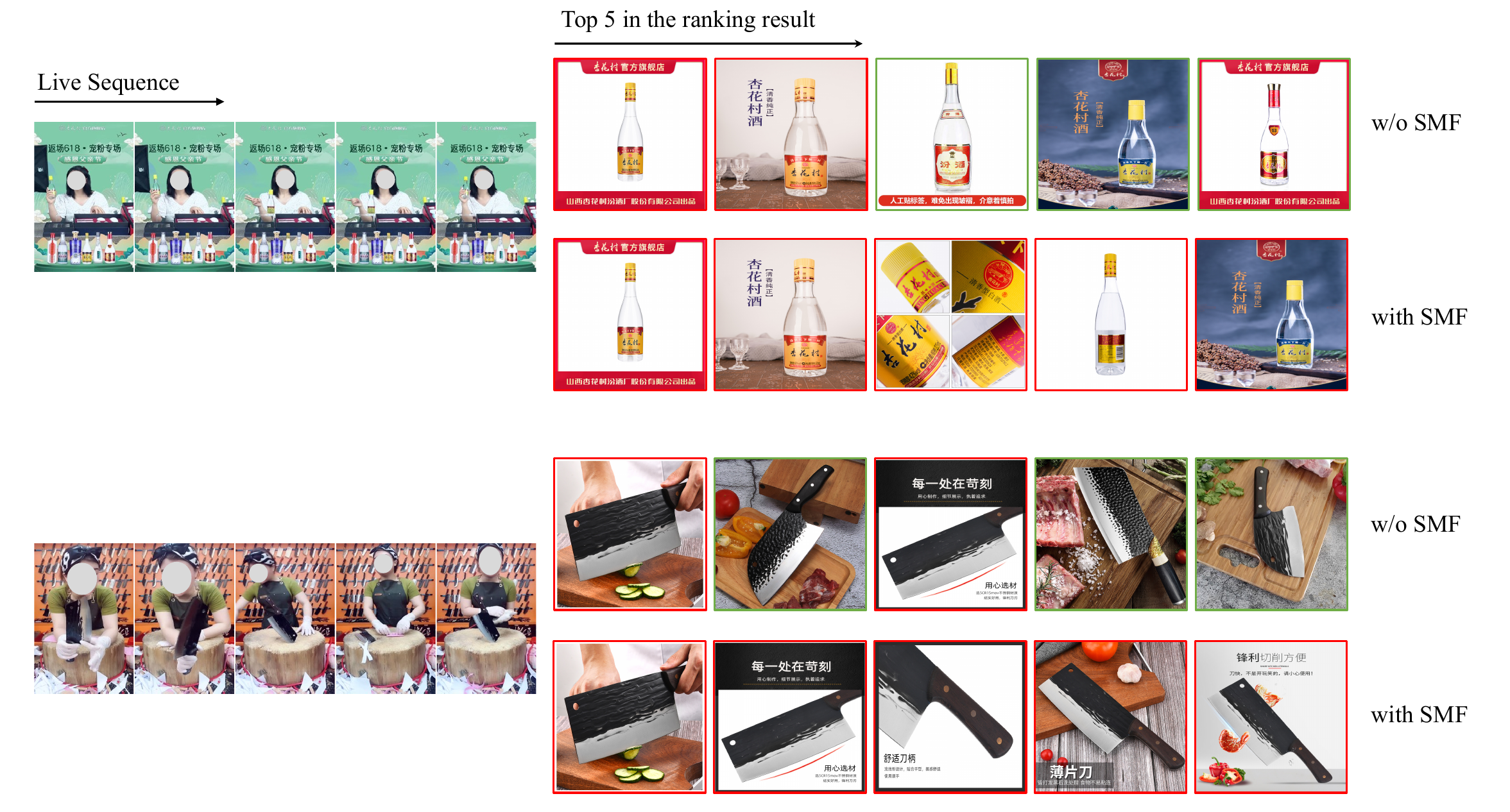}}
    \vspace{-1.5mm} 
    \caption{The ranking results of the LPR4M dataset. (a) The top 5 shop images in the ranking results of models with and without (w/o) the TE module. The texts below the video and image are ASR text and product titles, respectively. Images with red and green boundaries denote true positives and false positives. (b) The top 5 shop images in the ranking results of models with and without (w/o) the SMF module in highly similar background products. }
    \label{sup_rank_lpr4m}
\vspace{-5.5mm}    
\end{figure*}

\noindent\textbf{Ranking Results on the LPR4M Dataset.} We randomly select several video sequences in the LPR4M dataset and show the top 5 shop images in the ranking results of models with and without (w/o) the TE module in cluttered background products (Fig.~\ref{sup_background}) and models with and without (w/o) the SMF module in highly similar background products (Fig.~\ref{sup_similar}). The TE model guides the network to accurately retrieve the intended products related to the video ASR keywords among various distracting products. The SMF module assists our model in capturing sufficient fine-grained features to distinguish similar samples.

\begin{figure*}[t]
    \centering
    \includegraphics[width=0.95\linewidth]{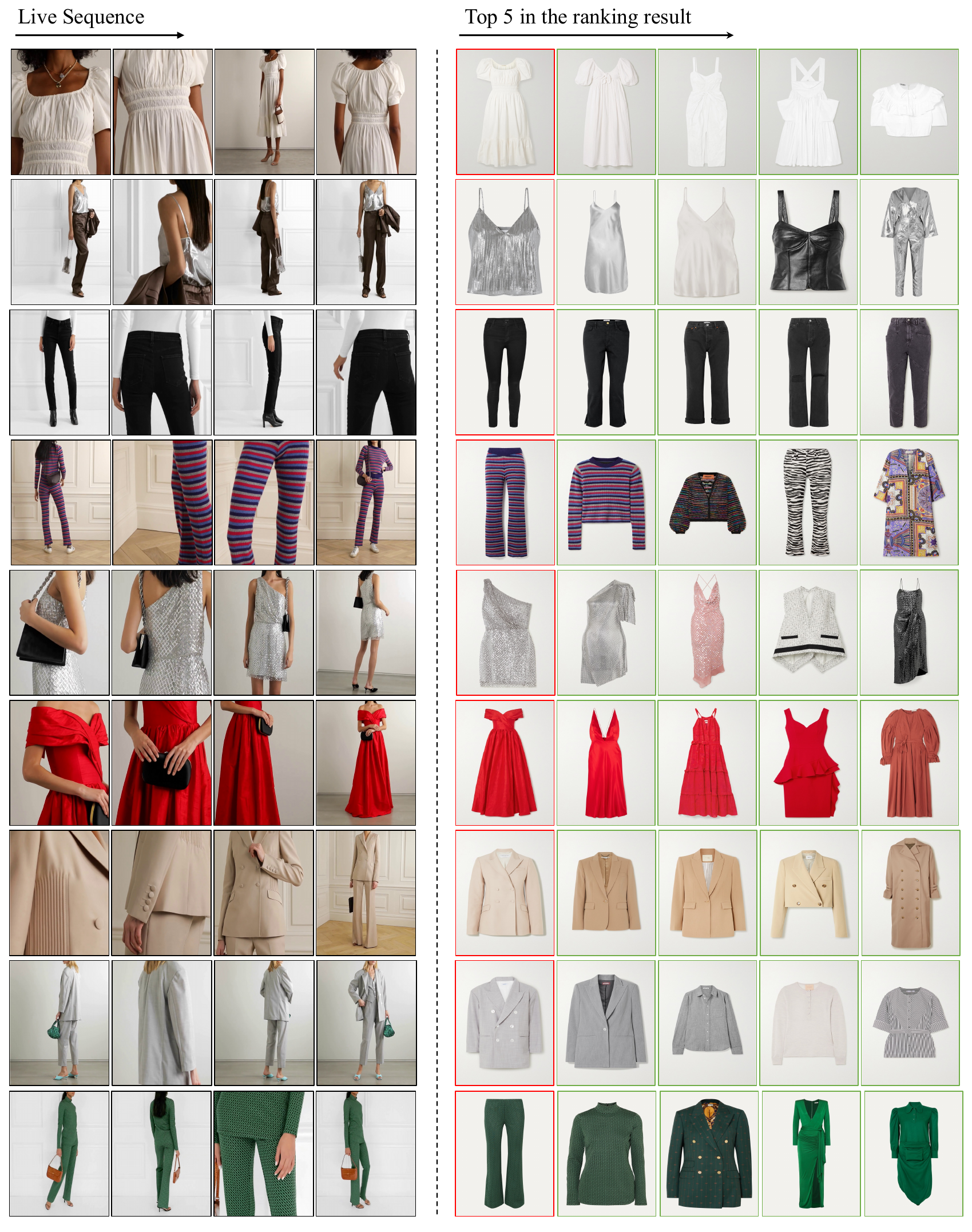}
    \vspace{-2.5mm} 
    \caption{The top 5 shop images in the ranking results of several live sequences in the MF dataset. Images with red and green boundaries denote true positives and false positives.}
    \label{sup_rank_mf}
\vspace{-3mm} 
\end{figure*}

\noindent\textbf{Ranking Results on the MF Dataset.} We randomly select several video sequences in the MF dataset and show the top 5 shop images in the ranking results in Fig.~\ref{sup_rank_mf}. It can be seen that even if the MF dataset does not have the text modal, our SGMN without text domain features can still accurately retrieve the products that best match the current video among a large number of clothing images.

\vspace{-2mm} 








\end{document}